\documentclass[sigconf,nonacm,balance=false]{acmart}
\setcopyright{none}
\let\originalcopyrightfootnote\footnotetextcopyrightpermission
\renewcommand{\footnotetextcopyrightpermission}[1]{%
  \originalcopyrightfootnote{\textit{Preprint.}}%
}
\AtBeginDocument{%
  }
\usepackage{algorithm}
\usepackage{algorithmic}
\usepackage{amsmath}
\usepackage{tabularx}
\usepackage{array}
\usepackage{graphicx}
\usepackage{booktabs}
\usepackage{colortbl}
\usepackage{float}
\usepackage{pifont}

\newcommand{\yesmark}{\ding{51}}
\newcommand{\nomark}{\ding{55}}

\newcommand{\DocSkillCompactAuthorBlock}{%
  \begin{minipage}{\textwidth}
    \centering
    {\normalfont\fontsize{11}{14}\selectfont
      \mbox{Jing Huang\textsuperscript{1}},
      \mbox{Zhiya Tan\textsuperscript{1,2}},
      \mbox{Changtao Miao\textsuperscript{1}},
      \mbox{Shutao Gong\textsuperscript{1}},\\[1pt]
      \mbox{Yang Wang\textsuperscript{1}},
      \mbox{Joey Tianyi Zhou\textsuperscript{3,4}},
      \mbox{Jianshu Li\textsuperscript{1}},
      \mbox{Ying Yan\textsuperscript{1}}\par
    }
    \vspace{3pt}
    {\normalfont\fontsize{9}{12}\selectfont
      \textsuperscript{1}Ant Digital Technologies, Ant Group\quad
      \textsuperscript{2}Nanyang Technological University, Singapore\par
      \textsuperscript{3}Singapore Management University, Singapore\par
      \textsuperscript{4}Institute of Advanced Intelligence and Computing (IAIC),\par
      Agency for Science, Technology and Research (A*STAR), Singapore\par
    }
  \end{minipage}%
}
\makeatletter
\renewcommand{\@mkauthors@iii}{%
  \global\setbox\mktitle@bx=\vbox{%
    \hsize=\textwidth
    \unvbox\mktitle@bx
    \par\medskip
    \noindent\DocSkillCompactAuthorBlock
    \par\bigskip
  }%
}
\makeatother

\begin{document}

\title{DocSkill: Skill-Integrated Agentic Reasoning for Multilingual Multi-Hop Document Question Answering}

\author{Jing Huang}
\email{jh.jj@antgroup.com}
\orcid{0009-0005-5651-9033}
\affiliation{%
  \institution{Ant Digital Technologies, Ant Group}
  \city{Singapore}
  \country{Singapore}
}

\author{Zhiya Tan}
\email{zhiya.tan@antgroup.com}
\affiliation{%
  \institution{Ant Digital Technologies, Ant Group}
  \city{Singapore}
  \country{Singapore}
}
\affiliation{%
  \institution{Nanyang Technological University}
  \city{Singapore}
  \country{Singapore}
}

\author{Changtao Miao}
\email{miaochangtao.mct@antgroup.com}
\orcid{0000-0002-7634-9992}
\affiliation{%
  \institution{Ant Digital Technologies, Ant Group}
  \city{Hangzhou}
  \country{China}
}

\author{Shutao Gong}
\email{gongshutao.gst@digital-engine.com}
\affiliation{%
  \institution{Ant Digital Technologies, Ant Group}
  \city{Changsha}
  \country{China}
}

\author{Yang Wang}
\email{wy462625@antgroup.com}
\affiliation{%
  \institution{Ant Digital Technologies, Ant Group}
  \city{Beijing}
  \country{China}
}

\author{Joey Tianyi Zhou}
\email{tyzhou@smu.edu.sg}
\affiliation{%
  \institution{Singapore Management University}
  \city{Singapore}
  \country{Singapore}
}

\affiliation{%
  \institution{Institute of Advanced Intelligence and Computing (IAIC), Agency for Science, Technology and Research (A*STAR)}
  \city{Singapore}
  \country{Singapore}
}

\author{Jianshu Li}
\email{jianshu.l@antgroup.com}
\affiliation{%
  \institution{Ant Digital Technologies, Ant Group}
  \city{Singapore}
  \country{Singapore}
}

\author{Ying Yan}
\email{fuying.yy@antgroup.com}
\affiliation{%
  \institution{Ant Digital Technologies, Ant Group}
  \city{Singapore}
  \country{Singapore}
}

\renewcommand{\shortauthors}{Jing Huang et al.}

\begin{abstract}
Multilingual multi-hop document question answering (MMHop-DocQA) requires locating and integrating evidence across complex
layouts, document pages, and diverse languages. Existing approaches include single-pass processing and adaptive tool use, but
learning to coordinate evidence acquisition, cross-page integration,
and answer verification remains challenging. To address this challenge, we present
\textbf{DocSkill}, \textit{an agentic framework with a learned
skill-orchestration policy for adaptive evidence acquisition
and reasoning}. The policy selects, revisits, or skips skills according
to the evolving evidence state. To construct supervision
for skill orchestration, bounded Monte Carlo Tree Search
(MCTS) explores alternative skill trajectories offline.
These trajectories provide demonstrations for supervised
fine-tuning (SFT) and preference pairs for Direct
Preference Optimization (DPO). We further introduce \textbf{MMHop-DocBench}, a multilingual
multi-hop document QA dataset with training and test splits,
containing 60K QA pairs across 10 languages and eight document
domains. Its \textit{HopGen} pipeline builds hierarchical
page--element graphs to derive traceable cross-page evidence
paths and generate grounded QA pairs with page- and
element-level evidence annotations. Experiments show that DocSkill with Qwen3-VL-8B achieves 41.0 accuracy and 49.2 $\mathrm{F}_1$ on MMHop-DocBench, exceeding the strongest evaluated baseline by 5.5 and 6.3 points, respectively,
and achieving competitive performance on
MMLongBench-Doc and DocBench.

\end{abstract}


\maketitle
\fancyhead[LE,RO]{}

\begin{figure}[t]
  \centering
  \includegraphics[width=\linewidth]{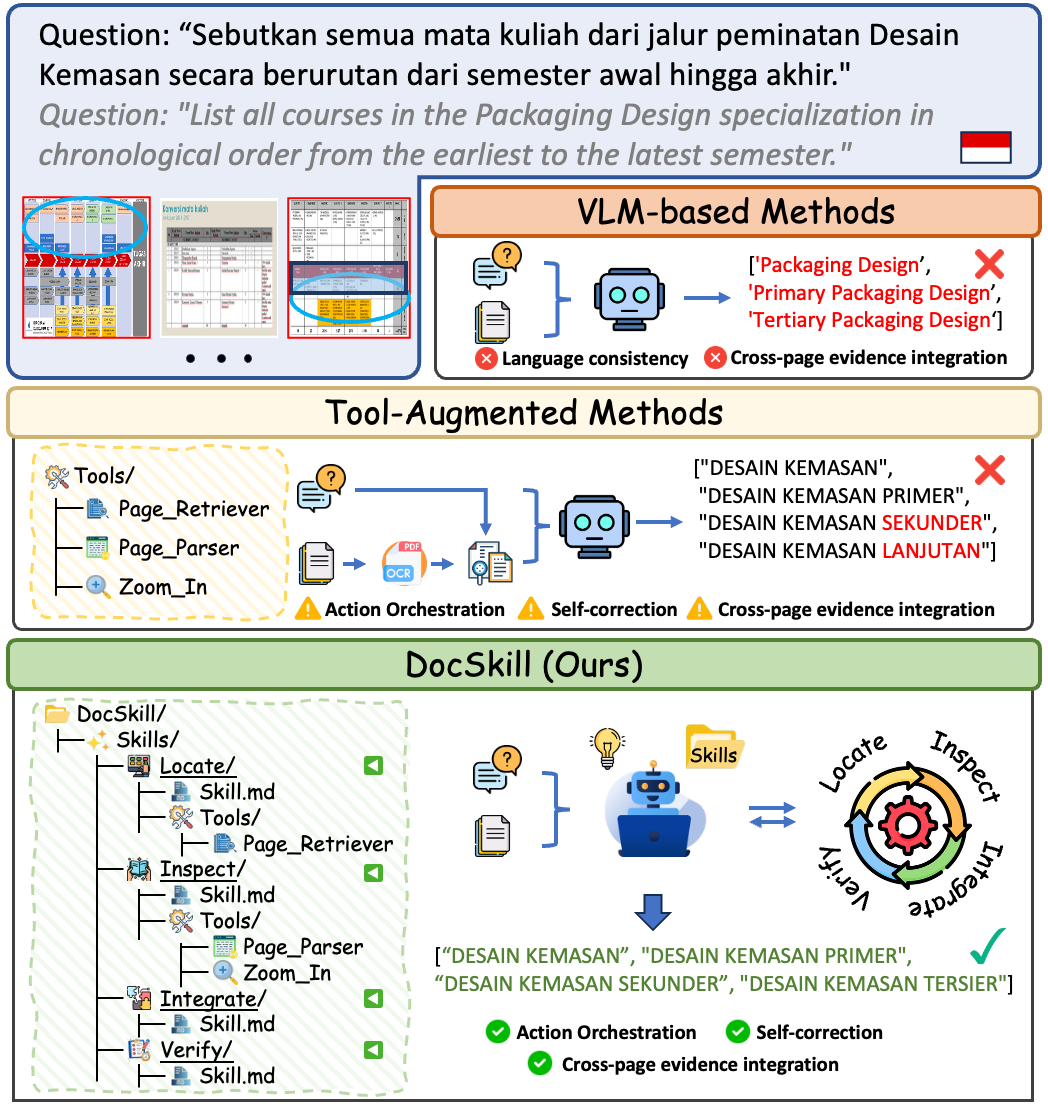}
  \caption{Overview of DocSkill and its comparison with VLM-based and tool-augmented methods. DocSkill adaptively orchestrates four skills, \textit{Locate}, \textit{Inspect}, \textit{Integrate}, and \textit{Verify}, for cross-page document reasoning and answering.}
  \label{fig:overview}
  \Description{Diagram comparing VLM-based methods,
  tool-augmented methods, and DocSkill. DocSkill adaptively
  selects four document skills to retrieve relevant pages,
  inspect fine-grained content, integrate cross-page
  evidence, and verify candidate answers.}
\end{figure}

\section{Introduction}
Multilingual multi-hop document question answering (MMHop-DocQA) is a cornerstone capability for intelligent systems that process multilingual, multi-page documents
\cite{mathew2021docvqadatasetvqadocument,
tito2023hierarchicalmultimodaltransformersmultipage,
ma2024mmlongbenchdocbenchmarkinglongcontextdocument,
zou2024docbenchbenchmarkevaluatingllmbased,
huang2025lavcotlanguageawarevisualcot,
tang2025mtvqabenchmarkingmultilingualtextcentric}. The fundamental challenge lies in locating and integrating sparse, spatially distributed evidence across heterogeneous layouts and languages, requiring iterative document navigation, structural grounding, and multi-step reasoning over long contexts.

As illustrated in Figure~\ref{fig:overview}, recent approaches can be broadly grouped into two paradigms with progressively increasing reasoning capabilities, yet each exhibiting inherent limitations. 
\textbf{(a) VLM-based Methods} directly encode document pages and queries into a Vision-Language Model (VLM)
\cite{bai2025qwen3vltechnicalreport, openai2026gpt56systemcard, anthropic2026claudesonnet5, googledeepmind2026gemini31pro}. Although these methods benefit from end-to-end multimodal understanding, single-pass inference over long document contexts can suffer from context overload, language inconsistency, and incomplete integration of sparse evidence across pages. \textbf{(b) Tool-Augmented Methods} enhance document processing through OCR \cite{cui2025paddleocr30technicalreport},
retrieval-augmented generation~\cite{yu2025visragvisionbasedretrievalaugmentedgeneration, faysse2025colpaliefficientdocumentretrieval, cho2024m3docragmultimodalretrievalneed, guo2025raganythingallinoneragframework},
or agentic tool use
\cite{han2025mdocagentmultimodalmultiagentframework, jain2025simpledocmultimodaldocumentunderstanding, sun2025docagent, zhang2026docdanceragenticdocumentgroundedinformation}. While these mechanisms improve evidence access and
support adaptive tool use, learning to jointly coordinate
evidence acquisition, cross-page integration, and answer
verification remains challenging.
These limitations are further amplified in multilingual settings, which require robust understanding across languages and reliable evidence grounding throughout the reasoning process.

To address these limitations, we introduce \textbf{DocSkill}, an agentic framework for learned skill orchestration in multilingual multi-hop document QA.
Its key design shifts agent control from operation-level tool selection to capability-level skill orchestration.
Specifically, we distinguish a \emph{skill} from a \emph{tool}: a skill is a reusable capability abstraction that specifies what the agent should accomplish given the current interaction state, whereas a tool is a concrete executable operation that specifies how part of that capability is carried out.
Accordingly, a skill may combine model-native reasoning with one or more tool calls, decoupling orchestration decisions from specific tool implementations. As illustrated in Figure~\ref{fig:overview}, DocSkill organizes document interaction into four reusable skills:
\textit{Locate} combines query refinement with page retrieval via Page\_Retriever;
\textit{Inspect} extracts and interprets fine-grained multimodal content, invoking Page\_Parser or Zoom\_In when needed;
\textit{Integrate} connects complementary findings across pages; and
\textit{Verify} checks the grounding and consistency of candidate answers against accumulated evidence.
Conditioned on the evolving interaction state, the learned policy adaptively selects, skips, repeats, and composes these skills.

Learning adaptive skill orchestration requires trajectory-level
supervision because the effectiveness of each skill decision
depends on its downstream contribution to evidence acquisition,
reasoning, and answer generation.
We adopt an offline trajectory curation and preference
optimization strategy.
We use bounded Monte Carlo Tree Search (MCTS)
\cite{silver2017masteringchessshogiselfplay,
yao2023treethoughtsdeliberateproblem}
to explore alternative trajectories over the skill space.
The resulting trajectories are ranked by answer correctness
and overall quality, assessed through language consistency,
logical soundness, and skill-orchestration effectiveness,
to construct training data for supervised fine-tuning (SFT)
and trajectory-level preference pairs.
We optimize the orchestration policy through SFT followed
by Direct Preference Optimization (DPO)
\cite{rafailov2024directpreferenceoptimizationlanguage,
xie2024montecarlotreesearch,
chen2024steplevelvaluepreferenceoptimization},
which learns from preferred and dispreferred skill trajectories.

To address the scarcity of multilingual multi-hop
document QA data, we introduce \textbf{MMHop-DocBench}
with train and test splits to support
policy learning and benchmark evaluation.
The dataset is constructed using \textit{HopGen},
an automatic curation pipeline that represents long documents
as hierarchical Page--Element Graphs
\cite{ho2020constructingmultihopqadataset,
xiao2019dynamicallyfusedgraphnetwork}
and derives traceable cross-page evidence paths
for grounded question generation.
MMHop-DocBench contains 60K question--answer pairs
across 10 languages and 8 document domains, together
with explicit page- and element-level evidence annotations.

Experiments show that DocSkill outperforms strong
baselines on MMHop-DocBench and achieves competitive
performance on both MMLongBench-Doc and DocBench.

Our main contributions are summarized as follows:
\begin{itemize}
    \item \textbf{Agentic Skill-Orchestration Framework.}
    We introduce DocSkill, which orchestrates four reusable
reasoning and tool-use skills. Bounded MCTS generates
trajectories for SFT and preference pairs for DPO.

    \item \textbf{Multilingual Multi-Hop QA Dataset.}
We introduce HopGen to construct MMHop-DocBench, comprising
60K QA pairs across 10 languages and eight domains with
page- and element-level evidence annotations.

    \item \textbf{Empirical Results and Analysis.}
Experiments on three benchmarks evaluate DocSkill against
strong baselines, with comparisons and ablations analyzing
skill orchestration and trajectory supervision.
\end{itemize}

\begin{figure*}[t]
  \centering
  \includegraphics[width=\textwidth]{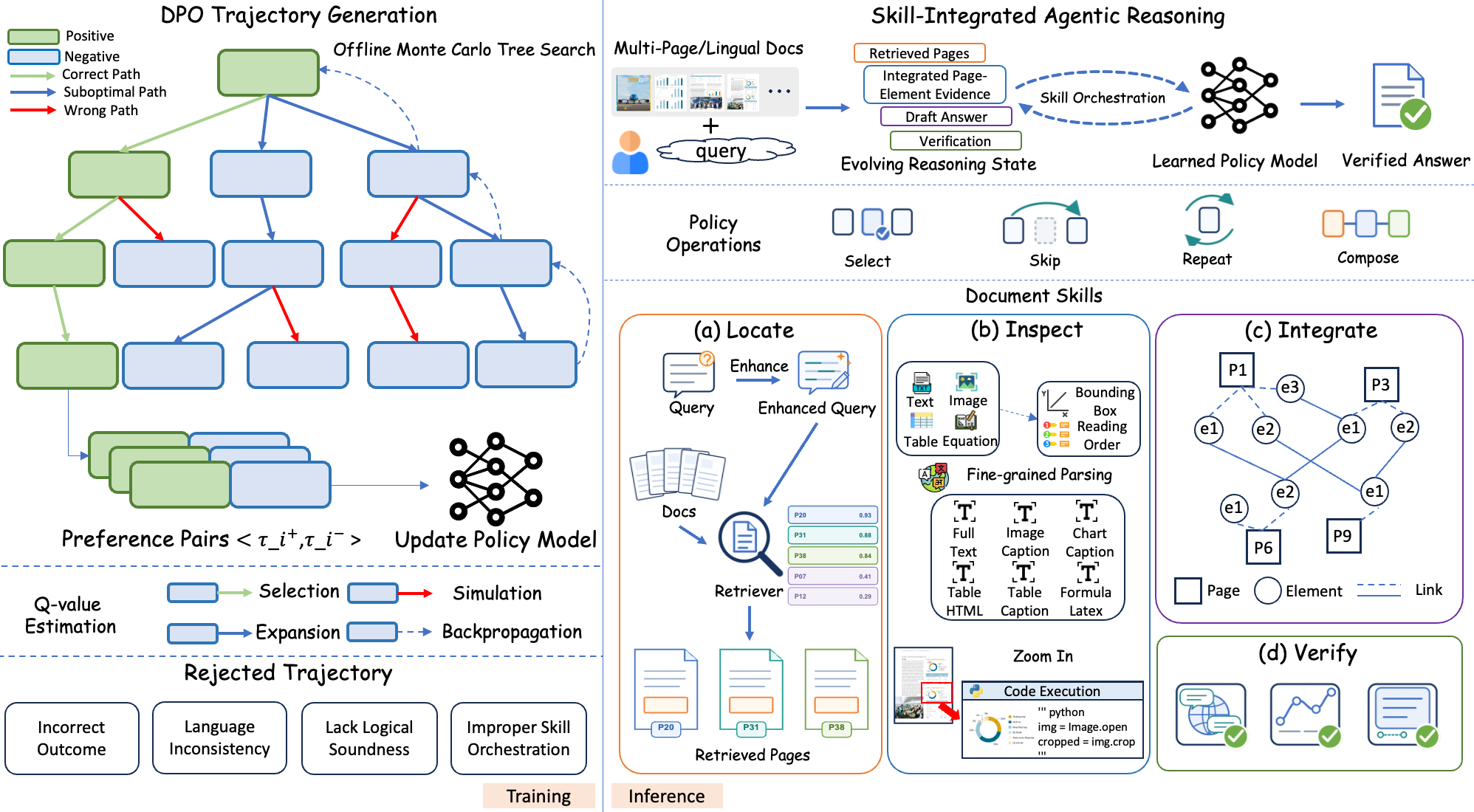}
  \caption{Overview of the DocSkill framework. During offline training, bounded MCTS explores alternative skill trajectories to construct training data for SFT and preference pairs for DPO. During inference, the learned policy adaptively orchestrates four skills, \textit{Locate}, \textit{Inspect}, \textit{Integrate}, and \textit{Verify}, each combining model reasoning with optional tool use.}
  \label{fig:framework}
  \Description{Framework diagram showing offline trajectory
  construction and inference-time skill orchestration.
  Bounded MCTS generates supervision for SFT and DPO.
  At inference, the learned policy selects document skills
  based on the evolving interaction state. Each skill
  performs model-native reasoning, optionally invokes
  tools, and updates the state for subsequent decisions.}
\end{figure*}

\section{Related Work}
\subsection{Document Question Answering}
Document Question Answering (DocQA) involves extracting
and reasoning over textual and visual evidence in
documents.
VLM-based approaches process document images directly
\cite{bai2025qwen3vltechnicalreport,
openai2026gpt56systemcard,
googledeepmind2026gemini31pro},
while OCR-based pipelines
\cite{cui2025paddleocr30technicalreport}
extract text and layout information for subsequent
reasoning.
Retrieval-augmented generation (RAG)
\cite{yu2025visragvisionbasedretrievalaugmentedgeneration,
faysse2025colpaliefficientdocumentretrieval,
cho2024m3docragmultimodalretrievalneed,
guo2025raganythingallinoneragframework}
selects relevant evidence from searchable document
indexes to support answer generation. Document agents extend these approaches through
iterative evidence acquisition and tool use
\cite{han2025mdocagentmultimodalmultiagentframework,
jain2025simpledocmultimodaldocumentunderstanding,
sun2025docagent,
zhu2025doclenstoolaugmentedmultiagent}.
DeepRead~\cite{li2026deepreaddocumentstructureawarereasoning}
supports structure-aware retrieval and section reading,
while ORCA~
\cite{lassoued2026orcaorchestratedreasoningcollaborative}
coordinates specialized agents through routing and debate.
Doc-$V^{*}$~
\cite{zheng2026docvcoarsetofineinteractivevisualreasoning}
combines OCR-free page navigation with working memory,
using SFT and GRPO for policy training. Our work focuses on multilingual multi-hop DocQA,
learning state-dependent orchestration over
capability-level skills through SFT and DPO
on offline MCTS trajectories.

\subsection{Skill-integrated Agentic Reasoning}
Skill learning for LLM agents focuses on acquiring structured
reasoning and action strategies. Prior work, including
ReAct~\cite{yao2023reactsynergizingreasoningacting} and
tool-oriented approaches such as
Toolformer~\cite{schick2023toolformerlanguagemodelsteach},
ToolAlpaca~\cite{tang2023toolalpacageneralizedtoollearning},
and Gorilla~\cite{patil2023gorillalargelanguagemodel},
demonstrates the effectiveness of interleaving reasoning
with tool use. These approaches establish reasoning and
tool-use interfaces. Some recent works have explored
skill-integrated agent frameworks, such as
AutoSkill~\cite{yang2026autoskillexperiencedrivenlifelonglearning}
and XSkill~\cite{jiang2026xskillcontinuallearningexperience},
where task-level skills and action strategies are explicitly
defined and composed to solve complex tasks.
In this work, we introduce DocSkill, a skill-integrated
agentic framework for multilingual multi-hop document
question answering that dynamically composes reasoning
actions and tool usage based on evolving interaction states
for iterative evidence acquisition and structured document reasoning.

\section{Method}
\label{sec:method}

Figure~\ref{fig:framework} summarizes DocSkill's offline
training and adaptive inference.
Conditioned on the interaction state, the policy selects
among \textit{Locate}, \textit{Inspect}, \textit{Integrate},
and \textit{Verify}, each combining reasoning with
optional tool execution.
To construct training data, bounded MCTS uses a frozen
generator and evaluator to produce scored skill trajectories
with explicit reasoning from MMHop-DocBench QA instances.
The resulting trajectories provide SFT demonstrations
and DPO preference pairs for policy training.
At inference, the learned policy orchestrates document
skills without tree search.

\subsection{Skill-Orchestrated Document Agent}
\label{sec:skill_agent}

Given an input $x=(\mathcal{D},q)$, where
$\mathcal{D}=\{p_i\}_{i=1}^{M}$ is an $M$-page multilingual
document and $q$ is a question, the goal is to generate an
answer $y$ grounded in cross-page evidence.
Following ReAct~\cite{yao2023reactsynergizingreasoningacting},
we formulate DocSkill as a finite-step interaction with
the document environment, in which the agent acquires
evidence, updates its reasoning state, and determines
its subsequent actions.

\subsubsection{Skill Abstraction}
In DocSkill, a skill is a reusable document-level capability
that organizes model reasoning and optional tool
execution around a specific objective. A tool, in contrast,
provides an executable operation, such as page retrieval,
document parsing, or region cropping.
This distinction concerns the abstraction used for agent
control: the policy selects the document capability to
invoke and specifies its arguments, while the corresponding
skill carries out the reasoning and tool execution needed
for that invocation.
Each skill is represented as
\begin{equation}
\sigma_k=\langle d_k,\mathcal{U}_k\rangle,
\label{eq:skill_definition}
\end{equation}
where $d_k$ describes the capability and
$\mathcal{U}_k$ denotes its available tool set, which may
be empty for reasoning-only execution.
The skill library is
\begin{equation}
\mathcal{S}
=
\{
\mathsf{Locate},
\mathsf{Inspect},
\mathsf{Integrate},
\mathsf{Verify}
\}.
\label{eq:skill_space}
\end{equation}
All four skills are predefined and fixed
during policy training.

\subsubsection{Interaction State and Policy}
At step $t$, the interaction state is
\begin{equation}
z_t=
\left(
q,\mathcal{E}_t,\hat{y}_t,v_t,\mathcal{H}_t
\right),
\label{eq:agent_state}
\end{equation}
where $q$ is the original input question;
$\mathcal{E}_t$ contains accumulated page- and
element-level evidence and connections across pages;
$\hat{y}_t$ denotes the current candidate answer;
$v_t$ records verification and reflection feedback; and $\mathcal{H}_t$ is the
ordered history of preceding actions and their
returned observations.
The input document $\mathcal{D}$ remains accessible
through the document environment and is consulted
via skill executions and their associated tools.

The policy operates over an action space $\mathcal{A}$
containing parameterized skill invocations and terminal
answer outputs:
\begin{equation}
a_t\sim\pi_{\theta}(\cdot\mid z_t),
\qquad
a_t\in\mathcal{A}.
\label{eq:skill_policy}
\end{equation}
A skill invocation takes the form $a_t=(s_t,\xi_t)$,
where $s_t\in\mathcal{S}$ specifies the skill and
$\xi_t$ contains its arguments, such as a retrieval query,
target pages, or regions of interest.
Thus, invoking the same skill with different arguments
constitutes different actions. Alternatively, the policy emits a terminal action, which returns the final
answer $y$ and ends the interaction.

\subsubsection{Document Skills}
The four skills perform complementary operations on
the interaction state.

\textit{Locate} uses model reasoning to formulate
or refine a query based on the question and interaction
state, then invokes \texttt{Page\_Retriever} to identify
candidate evidence pages. Returned page references and
content are added to the accumulated evidence.
The policy can revisit this skill when further evidence
is needed.

\textit{Inspect} invokes \texttt{Page\_Parser} for
layout-aware extraction and, when needed,
\texttt{Zoom\_In} for region cropping.
Model-native reasoning interprets the returned content,
and the resulting element-level findings are added
to the accumulated evidence.

\textit{Integrate} uses model-native reasoning to connect
facts across pages and synthesize accumulated evidence
to construct or refine the candidate answer, without
additional retrieval or extraction.

\textit{Verify} uses model reasoning to assess
the candidate answer for completeness, consistency,
and evidence support. Its feedback guides further
evidence acquisition and answer revision.

\subsubsection{Execution and Composition.}
Executing a skill returns an observation, updates the
relevant state components, and records the action and
observation in the interaction history.
Trajectories retain the agent's recorded reasoning,
skill invocations, and native skill outputs, including
candidate answers and verification feedback.
Native skill outputs serve as observations that
condition subsequent decisions.
Tools are invoked within skills rather than exposed
as independent top-level actions. Skills compose sequentially through the shared state.
At each step, the policy selects a skill invocation
or a terminal answer. The updated state allows skills
to be revisited or skipped without a prescribed order.
These sequences are selected dynamically rather than
defined as separate compound skills.

A completed interaction with $T$ skill invocations is
denoted as
\begin{equation}
\tau=
(z_0,a_0,o_1,z_1,\ldots,a_{T-1},o_T,z_T,y),
\label{eq:trajectory}
\end{equation}
where $a_t$ denotes a skill invocation,
$o_1,\ldots,o_T$ are the observations,
and $y$ is the final answer.
Incomplete rollouts retain the recorded interaction
prefix without a terminal answer.
These trajectories define the space explored during
offline supervision construction.

\subsection{Offline MCTS-Based Skill Trajectory Curation}
\label{sec:mcts_trajectory}

The parameterized interactions defined above induce a
large trajectory space, where the value of a skill
invocation may depend on its downstream effects on
evidence acquisition, integration, and verification.
Independent rollouts can repeatedly explore similar
paths and provide limited coverage of alternative
decisions.
We therefore apply bounded Monte Carlo Tree Search
(MCTS)~\cite{silver2017masteringchessshogiselfplay,
yao2023treethoughtsdeliberateproblem} to allocate a
fixed search budget across candidate skill trajectories.

For each training instance
$x_i=(\mathcal{D}_i,q_i)$, the root node $n_0$
represents the initial state $z_0$, and each subsequent
node represents an interaction state.
An edge records a complete action.
For a skill invocation $a_t=(s_t,\xi_t)$, executing the
edge produces an observation $o_{t+1}$ and transitions
the agent from $z_t$ to $z_{t+1}$.
A terminal answer action ends the corresponding
branch. 
Search depth counts policy decisions, with each
edge representing one skill invocation or terminal answer.
Each edge $e$ maintains an estimated value $Q(e)$
and a visit count $N(e)$, while each node $n$
maintains a visit count $N(n)$.

\paragraph{Selection.}
Starting from the root, MCTS recursively selects the
edge with the highest Upper Confidence Bound for Trees
(UCT) score:
\begin{equation}
\operatorname{UCT}(e)
=
Q(e)
+
\lambda
\sqrt{
\frac{\ln\!\left(N(n)+1\right)}
     {N(e)+1}
},
\label{eq:uct}
\end{equation}
where $n$ is the parent node of $e$, $Q(e)$ is the
mean return of previous rollouts through the edge, and
$\lambda$ balances exploration and exploitation.
Selection stops at an unexpanded or terminal node.

\paragraph{Expansion.}
At an expandable leaf, the generator proposes feasible
actions conditioned on the current state.
Skill invocations with different arguments form distinct
branches, producing observations and child states through
model reasoning and optional tool calls when executed.
Answer actions create terminal children.

\paragraph{Simulation.}
From an expanded child, the generator proposes and
executes actions until it emits a terminal answer,
reaches the maximum depth, or fails to acquire new
page- or element-level evidence after an
evidence-acquisition action.
This evidence-stagnation criterion is used to
truncate offline rollouts and is not an inference-time
stopping rule.
The evaluator assigns the trajectory quality score:
\begin{equation}
C(\tau)=
w_{\mathrm{lang}}c_{\mathrm{lang}}(\tau)
+w_{\mathrm{logic}}c_{\mathrm{logic}}(\tau)
+w_{\mathrm{skill}}c_{\mathrm{skill}}(\tau),
\label{eq:trajectory_quality}
\end{equation}
where the three components measure language consistency,
logical soundness, and skill-orchestration effectiveness,
respectively.
Each component takes a value in $\{0,0.5,1\}$.
The weights are non-negative and sum to one; we assign
equal weight to the three components.

The evaluator assigns an outcome score
$O(\tau)\in\{1,-1,0\}$ for correct, incorrect, and
incomplete trajectories, respectively.
The rollout return combines the outcome score
with trajectory quality:
\begin{equation}
J(\tau)=O(\tau)+C(\tau).
\label{eq:rollout_return}
\end{equation}

\paragraph{Backpropagation.}
The rollout return $J(\tau)$ is propagated through the
selected path.
For each visited node $n_j$ and edge $e_j$, we update
\begin{equation}
N(n_j)\leftarrow N(n_j)+1,
\label{eq:node_visit_update}
\end{equation}
\begin{equation}
N(e_j)\leftarrow N(e_j)+1,
\label{eq:edge_visit_update}
\end{equation}
and
\begin{equation}
Q(e_j)\leftarrow
Q(e_j)+
\frac{J(\tau)-Q(e_j)}{N(e_j)}.
\label{eq:edge_value_update}
\end{equation}
Thus, $Q(e_j)$ maintains the empirical mean return of
rollouts passing through the edge.

We use up to $B=3$ candidate actions per expansion,
a maximum search depth of $D=6$, and
$N_{\mathrm{sim}}=8$ simulations per question.
Both the generator and evaluator use frozen
GPT-5.6-SOL~\cite{openai2026gpt56systemcard} with
role-specific prompts and separate inference contexts.
The generator proposes actions and performs model-native
skill reasoning, while reference answers and evidence
annotations are used only for offline evaluation.
For each question, the search returns scored trajectories
containing recorded explicit reasoning, skill invocations,
and execution outputs.
Successful trajectories provide SFT demonstrations,
while trajectories with different scores for the same
question form DPO preference pairs.

\subsection{Trajectory-Level Preference Learning}
\label{sec:preference_learning}

The scored trajectories produced by MCTS provide
complementary supervision for imitation and preference
learning.

\paragraph{Supervised fine-tuning.}
We supervise successful trajectories, including
reasoning, skill invocations, native skill outputs,
and terminal answers.
External tool returns and deterministic state updates
serve as context.
Let $\mathcal{D}_{\mathrm{SFT}}$ denote these
input--trajectory pairs and $u_{1:L_\tau}$ the
serialized tokens of $\tau$.
We set $m_j=1$ for model-generated outputs and
$m_j=0$ for context tokens.
The SFT objective is
\begin{equation}
\mathcal{L}_{\mathrm{SFT}}
=
-\mathbb{E}_{(x,\tau)\sim\mathcal{D}_{\mathrm{SFT}}}
\left[
\sum_{j=1}^{L_\tau}
m_j\log p_\theta(u_j\mid c_j)
\right],
\label{eq:sft}
\end{equation}
where $p_\theta$ is the shared backbone's token
distribution and $c_j$ contains the role-specific
prompt, available interaction context, and preceding
output tokens for the corresponding generation call.

\paragraph{Preference construction.}
For each input $x_i$, two generated trajectories form
an ordered preference pair when one receives a higher
rollout score than the other:
\begin{equation}
\mathcal{D}_{\mathrm{pref}}
=
\left\{
(x_i,\tau_i^{+},\tau_i^{-})
\;\middle|\;
J(\tau_i^{+})>J(\tau_i^{-})
\right\}.
\label{eq:preference_dataset}
\end{equation}
Both trajectories in each pair are generated for the
same question, and equal-score pairs are discarded.

\paragraph{Trajectory-level DPO}
Using the same supervision mask, we define the
conditional log-likelihood of the recorded
model-generated content as
\begin{equation}
\ell_\theta(x,\tau)
=
\sum_{j=1}^{L_\tau}
m_j\log p_\theta(u_j\mid c_j).
\label{eq:trajectory_likelihood}
\end{equation}
Terminal-answer tokens contribute only when present,
so this definition also applies to incomplete
trajectory prefixes.
We optimize trajectory preferences using
\begin{equation}
\begin{aligned}
\mathcal{L}_{\mathrm{DPO}}
= -\mathbb{E}_{(x,\tau^+,\tau^-)\sim
\mathcal{D}_{\mathrm{pref}}}
\Big[
\log \sigma\big(
\beta[
r_\theta(x,\tau^+)
-r_\theta(x,\tau^-)
]
\big)
\Big],
\end{aligned}
\label{eq:dpo}
\end{equation}
where
$r_\theta(x,\tau)=
\ell_\theta(x,\tau)-\ell_{\mathrm{ref}}(x,\tau)$,
$\sigma$ is the logistic sigmoid, and the reference
model is the frozen SFT checkpoint.
Both models use the same supervision mask and
conditioning contexts.

\begin{table}[t]
\centering
\small
\setlength{\tabcolsep}{2pt}
\renewcommand{\arraystretch}{0.95}
\resizebox{\columnwidth}{!}{%
\begin{tabular}{@{}lccccc@{}}
\toprule
\textbf{Benchmark}
& \textbf{\#Langs.}
& \textbf{M-Hop}
& \textbf{\#QA}
& \textbf{Avg. Pages}
& \textbf{Source} \\
\midrule

ChartQA~\cite{masry2022chartqabenchmarkquestionanswering}
& 1 & \nomark & 32.7K & 1.0 & Charts \\

EST-VQA~\cite{wang2020generalvalueevidencebilingual}
& 2 & \nomark & 28.2K & 1.0 & Scenes \\

MTVQA~\cite{tang2025mtvqabenchmarkingmultilingualtextcentric}
& 9 & \nomark & 28.6K & 1.0 & Multi \\

DocVQA~\cite{mathew2021docvqadatasetvqadocument}
& 1 & \nomark & 50.0K & 1.0 & Multi \\

MP-DocVQA~\cite{tito2023hierarchicalmultimodaltransformersmultipage}
& 1 & \nomark & 46.2K & 8.2 & Multi \\

SlideVQA~\cite{tanaka2023slidevqadatasetdocumentvisual}
& 1 & \yesmark & 14.5K & 20.0 & Slides \\

MMLongBench-Doc~\cite{ma2024mmlongbenchdocbenchmarkinglongcontextdocument}
& 1 & \yesmark & 1.1K & 47.5 & Multi \\

DocBench~\cite{zou2024docbenchbenchmarkevaluatingllmbased}
& 1 & \nomark & 1.1K & 66.0 & Multi \\
\midrule

\textbf{MMHop-DocBench}
& \textbf{10}
& \yesmark
& \textbf{60.0K}
& \textbf{68.2}
& \textbf{Multi} \\

\bottomrule
\end{tabular}%
}
\caption{Comparison of document QA datasets by language count,
multi-hop support, number of QA pairs, average document length,
and source.}
\label{tab:benchmarks}
\end{table}

\section{MMHop-DocBench}
\label{sec:mmhop_docbench}

We introduce MMHop-DocBench, a large-scale benchmark
for multilingual multi-hop document question answering.
It contains long documents from diverse domains and
provides page- and element-level evidence references for
each question--answer pair.
The benchmark is constructed using \textit{HopGen}, an
automatic pipeline that generates grounded questions from
traceable cross-page evidence paths.

\begin{figure}[t]
  \centering
  \includegraphics[width=0.9\linewidth]{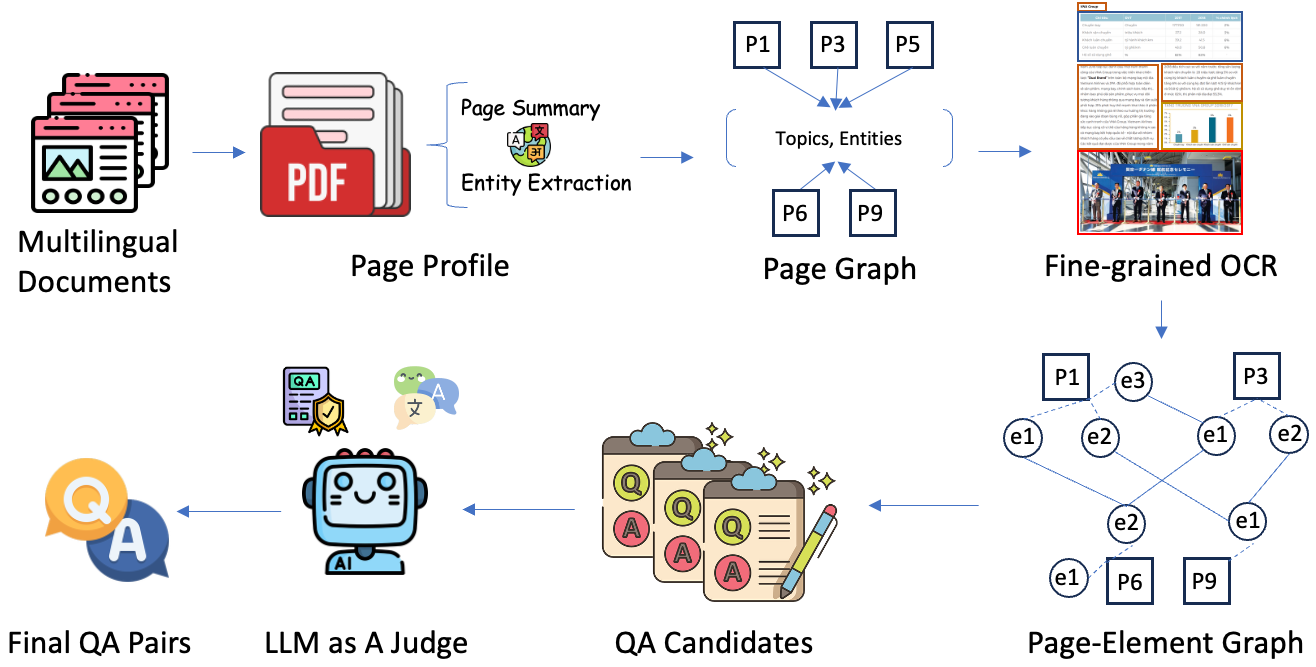}
  \caption{Overview of HopGen, which constructs hierarchical page--element graphs to generate and verify multilingual multi-hop QA pairs with explicit evidence references.}
  \label{fig:data_generation}
  \Description{HopGen pipeline for multilingual multi-hop
  QA curation. Page profiles and parsed elements form a
  hierarchical graph whose cross-page evidence paths guide
  question--answer generation. Candidates are evaluated,
  filtered, and deduplicated, with retained pairs linked
  to supporting pages and elements.}
\end{figure}

\subsection{HopGen: MMHop-DocQA Pair Curation}
\label{sec:hopgen}

Figure~\ref{fig:data_generation} provides an overview of
\textit{HopGen}.
HopGen uses GPT-5.6-SOL~
\cite{openai2026gpt56systemcard} for both generation and
evaluation, with role-specific prompts and separate inference
contexts, while PP-OCR~
\cite{cui2025paddleocr30technicalreport} provides
layout-aware parsing to extract fine-grained elements
from document pages.

For each document, the generator creates a compact profile
of every page, including its summary, entities, topics, and
layout structures.
It then assesses candidate page pairs for explicit semantic
dependencies to construct a Page Graph.
The parser extracts fine-grained document elements and
attaches them to their corresponding source pages,
forming a hierarchical Page--Element Graph.
Connected page sequences and their associated elements
define candidate cross-page evidence paths for grounded
question generation and subsequent verification. Given an evidence path and parsed elements, the generator
produces questions, answers, and supporting page--element
references in the document language using only the
supplied evidence. The generator is instructed to create questions requiring
complementary evidence from at least two distinct pages,
with no single page sufficient for the complete answer.
The evaluator scores candidates for answer correctness,
evidence sufficiency, cross-page necessity, and language
consistency, retaining those meeting a predefined threshold.
Normalized question--answer duplicates are resolved by
retaining the higher-scoring candidate. 

\begin{table*}[t]
\centering
\scriptsize
\setlength{\tabcolsep}{2pt}
\renewcommand{\arraystretch}{0.9}

\resizebox{\textwidth}{!}{%
\begin{tabular}{@{}l|*{10}{c}|cc@{}}
\toprule
\textbf{Method}
& \textbf{EN} & \textbf{ZH} & \textbf{YUE}
& \textbf{ID} & \textbf{MS} & \textbf{TL}
& \textbf{TA} & \textbf{TH} & \textbf{VI}
& \textbf{MY} & \textbf{Acc.} & $\boldsymbol{\mathrm{F}_1}$\\
\midrule
\multicolumn{13}{c}{\textbf{\textit{Vanilla VLM Methods}}} \\
\midrule

\textbf{Naive VLM} w/ Qwen3-VL-4B\cite{bai2025qwen3vltechnicalreport}
& 18.0/23.2 & 15.2/20.3 & 8.3/12.0
& 16.9/23.6 & 14.9/19.6 & 9.0/12.9
& 2.0/3.3 & 5.4/8.0 & 10.3/14.0
& 2.9/4.2 & 10.3 & 14.1 \\

\textbf{Naive VLM} w/ Qwen3-VL-8B\cite{bai2025qwen3vltechnicalreport}
& 24.0/28.9 & 21.0/25.9 & 12.0/15.7
& 20.6/28.0 & 19.8/23.9 & 13.5/16.6
& 3.6/5.1 & 8.2/10.5 & 14.3/18.2
& 4.3/5.7 & 14.1 & 17.9 \\

\textbf{Naive VLM} w/ Qwen3-VL-32B\cite{bai2025qwen3vltechnicalreport}
& 24.2/30.0 & 21.2/26.1 & 15.2/19.9
& 25.5/32.7 & 22.5/27.8 & 15.8/20.8
& 5.2/7.6 & 12.6/16.1 & 18.2/23.1
& 5.6/7.5 & 16.6 & 21.2 \\

\textbf{Naive VLM} w/ Gemini-3.1-Pro\cite{googledeepmind2026gemini31pro}
& 25.0/31.3 & \textbf{24.3}/29.8 & 20.8/27.5
& 25.4/33.7 & 23.8/30.5 & 20.4/28.6
& 12.9/18.6 & \textbf{21.8}/28.2 & 22.0/29.3
& 6.1/11.8 & 20.3 & 26.9 \\

\textbf{Naive VLM} w/ Claude-Sonnet-5\cite{anthropic2026claudesonnet5}
& 25.6/32.0 & 23.8/29.5
& \textbf{22.1/29.2}
& 26.0/34.5
& \textbf{24.8/31.4}
& \textbf{22.5/31.8}
& 13.5/19.8 & 20.8/28.1 & 22.4/29.8
& 6.4/12.5 & 20.8 & 27.9 \\

\textbf{Naive VLM} w/ GPT-5.6-SOL\cite{openai2026gpt56systemcard}
& \textbf{26.2/32.6}
& 24.0/\textbf{30.1}
& 21.8/28.8
& \textbf{26.5/35.2}
& 24.5/31.2 & 22.2/31.5
& \textbf{14.6/20.5}
& 21.4/\textbf{28.4}
& \textbf{22.7/30.1}
& \textbf{6.7/12.9}
& \textbf{21.1}
& \textbf{28.1} \\

\midrule
\multicolumn{13}{c}{\textbf{\textit{OCR-Based Methods}}} \\
\midrule

\textbf{Tesseract}\cite{smith2007overview} w/ GPT-5.6-SOL\cite{openai2026gpt56systemcard}
& 23.2/30.1 & 21.5/27.6 & 14.8/20.5
& 23.9/31.2 & 22.5/28.8 & 19.8/26.9
& 11.8/16.5 & 17.5/23.4 & 18.9/25.6
& 7.4/11.6 & 18.1 & 24.2 \\

\textbf{PP-OCR}\cite{cui2025paddleocr30technicalreport} w/ GPT-5.6-SOL\cite{openai2026gpt56systemcard}
& \textbf{27.1/33.5}
& \textbf{24.8/31.2}
& \textbf{21.5/28.4}
& \textbf{27.3/36.1}
& \textbf{25.4/32.3}
& \textbf{22.8/32.1}
& \textbf{14.2/20.1}
& \textbf{22.3/29.5}
& \textbf{23.1/30.8}
& \textbf{7.6/13.8}
& \textbf{21.6}
& \textbf{28.8} \\

\midrule
\multicolumn{13}{c}{\textbf{\textit{RAG-Based Methods}}} \\
\midrule

\textbf{VisRAG}~\cite{yu2025visragvisionbasedretrievalaugmentedgeneration} w/ GPT-5.6-SOL\cite{openai2026gpt56systemcard}
& 28.5/35.2 & 25.8/32.4 & 22.1/29.1
& 27.8/36.5 & 25.9/33.1 & 23.2/32.6
& 14.5/20.3 & 22.6/29.8 & 23.5/31.2
& 7.8/14.1 & 22.2 & 29.4 \\

\textbf{ColPali}~\cite{faysse2025colpaliefficientdocumentretrieval} w/ GPT-5.6-SOL\cite{openai2026gpt56systemcard}
& 29.2/36.1 & 26.4/33.2
& \textbf{23.1}/29.8
& 28.5/37.4 & 26.5/34.0 & 23.8/33.2
& 14.8/20.8 & 23.1/30.4 & 24.0/31.8
& 8.0/14.5 & 22.7 & 30.1 \\

\textbf{Qwen3-VL-Emb-8B}~\cite{li2026qwen3vlembeddingqwen3vlrerankerunifiedframework} w/ GPT-5.6-SOL\cite{openai2026gpt56systemcard}
& \textbf{30.8/38.5}
& \textbf{28.2/35.6}
& 22.6/\textbf{32.1}
& \textbf{30.1/39.2}
& \textbf{28.1/36.2}
& \textbf{25.4/35.1}
& \textbf{16.2/22.6}
& \textbf{24.8/32.5}
& \textbf{25.6/33.9}
& \textbf{8.9/15.8}
& \textbf{24.1}
& \textbf{32.2} \\

\midrule
\multicolumn{13}{c}{\textbf{\textit{Agent-Based Methods}}} \\
\midrule

\textbf{DocAgent}~\cite{sun2025docagent} w/ GPT-5.6-SOL\cite{openai2026gpt56systemcard}
& 27.9/34.2 & 25.5/31.9 & 22.2/29.0
& 28.0/36.7 & 26.1/32.9 & 23.5/32.7
& 14.9/20.8 & 23.0/30.1 & 23.8/31.4
& 8.2/14.4 & 22.3 & 29.4 \\

\textbf{SimpleDoc}~\cite{jain2025simpledocmultimodaldocumentunderstanding} w/ GPT-5.6-SOL\cite{openai2026gpt56systemcard}
& 28.6/35.0 & 26.2/32.8 & 22.8/29.9
& 28.7/37.6 & 26.8/33.8 & 24.2/33.6
& 15.5/21.6 & 23.6/30.9 & 24.4/32.2
& 8.6/15.0 & 22.9 & 30.2 \\

\textbf{DocLens}~\cite{zhu2025doclenstoolaugmentedmultiagent} w/ Gemini-3.1-Pro~\cite{googledeepmind2026gemini31pro}
& 38.6/45.5 & 36.3/43.2 & 33.6/40.9
& 38.3/45.6 & 36.9/44.2 & 34.7/42.2
& 25.9/32.6 & 34.3/41.6 & 35.2/42.9
& 18.6/25.2 & 33.2 & 40.4 \\

\textbf{DocLens}~\cite{zhu2025doclenstoolaugmentedmultiagent} w/ Claude-Sonnet-5~\cite{anthropic2026claudesonnet5}
& 40.3/47.4 & 37.6/44.7 & 34.7/42.0
& 39.6/47.0 & 38.0/45.5 & 36.3/44.1
& 27.3/34.3 & 35.6/43.1 & 36.7/44.6
& 20.3/27.0 & 34.6 & 42.0 \\

\textbf{DocLens}~\cite{zhu2025doclenstoolaugmentedmultiagent} w/ GPT-5.6-SOL~\cite{openai2026gpt56systemcard}
& 41.0/48.3 & 38.3/45.6 & 34.5/41.7
& 41.4/49.0 & 39.1/46.7 & 37.3/45.1
& 28.1/35.3 & 36.3/43.7 & 37.6/45.6
& 21.1/27.9 & 35.5 & 42.9 \\

\rowcolor[gray]{0.90}
\multicolumn{13}{l}{\textbf{DocSkill (Ours)}} \\

\quad w/ Gemini-3.1-Pro\cite{googledeepmind2026gemini31pro}
& 41.2/48.5 & 38.5/45.8 & 35.8/43.2
& 40.8/48.2 & 39.1/46.5 & 37.2/44.8
& 28.5/35.6 & 36.8/44.2 & 37.5/45.6
& 21.2/28.5 & 35.7 & 43.1 \\

\quad w/ Claude-Sonnet-5\cite{anthropic2026claudesonnet5}
& 42.5/49.8 & 39.8/47.2 & 36.5/44.1
& 42.1/49.5 & 40.5/47.8 & 38.5/46.2
& 29.8/37.2 & 37.8/45.5 & 38.6/46.8
& 22.5/30.1 & 36.9 & 44.4 \\

\quad w/ GPT-5.6-SOL\cite{openai2026gpt56systemcard}
& 43.8/51.2 & 41.2/48.6 & 36.2/43.8
& 43.5/51.0 & 41.8/49.2 & 39.8/47.8
& 31.2/38.8 & 37.5/45.2 & 39.8/48.2
& 23.8/31.6 & 37.9 & 45.5 \\

\quad w/ Qwen3-VL-4B \cite{bai2025qwen3vltechnicalreport} (SFT)
& 42.5/49.8 & 39.5/46.8 & 34.2/41.8
& 41.8/49.2 & 40.2/47.5 & 37.8/45.5
& 28.5/35.8 & 36.2/43.8 & 38.2/46.5
& 21.5/29.2 & 36.0 & 43.6 \\

\quad w/ Qwen3-VL-4B \cite{bai2025qwen3vltechnicalreport} (SFT + DPO)
& 44.2/51.8 & 41.5/49.2 & 35.8/43.5
& 43.8/51.5 & 42.5/50.2 & 39.5/47.8
& 30.8/38.5 & 38.2/46.2 & 40.2/48.8
& 23.5/31.8 & 38.0 & 45.9 \\

\quad w/ Qwen3-VL-8B \cite{bai2025qwen3vltechnicalreport} (SFT)
& 45.2/52.8 & 42.8/50.5 & 37.5/45.2
& 44.8/52.5 & 43.5/51.2 & 40.8/49.2
& 32.5/40.2 & 39.5/47.8 & 41.5/50.2
& 25.2/33.5 & 39.3 & 47.3 \\

\quad w/ Qwen3-VL-8B \cite{bai2025qwen3vltechnicalreport} (SFT + DPO)
& \underline{\textbf{46.8/54.5}}
& \underline{\textbf{44.5/52.2}}
& \underline{\textbf{39.2/47.1}}
& \underline{\textbf{46.2/54.2}}
& \underline{\textbf{45.2/53.1}}
& \underline{\textbf{42.5/51.2}}
& \underline{\textbf{34.2/42.5}}
& \underline{\textbf{41.2/49.5}}
& \underline{\textbf{43.2/52.1}}
& \underline{\textbf{27.1/35.8}}
& \underline{\textbf{41.0}}
& \underline{\textbf{49.2}} \\

\midrule
\multicolumn{13}{c}{\textbf{\textit{Human Baseline}}} \\
\midrule

\textbf{Human Experts}
& -- & -- & -- & -- & -- & --
& -- & -- & -- & -- & 61.7 & 63.9 \\

\bottomrule
\end{tabular}%
}
\caption{Performance on MMHop-DocBench across ten languages,
reported as per-language and macro-averaged Acc/$\boldsymbol{\mathrm{F}_1}$.
Bold values indicate the best results within each method
category, while underlined bold values denote the best model
results overall.}
\label{tab:vlm_baselines}
\end{table*}

\begin{table}[!t]
\centering
\scriptsize
\setlength{\tabcolsep}{2.5pt}
\renewcommand{\arraystretch}{1.0}

\resizebox{\columnwidth}{!}{%
\begin{tabular}{@{}l|ccc|c@{}}
\toprule
\textbf{Method}
& \multicolumn{3}{c|}{\textbf{MMLongBench-Doc}}
& \textbf{DocBench} \\
& \textbf{Acc.} & $\boldsymbol{\mathrm{F}_1}$ & \textbf{LASJ}
& \textbf{LASJ} \\
\midrule

\textbf{DocAgent}~\cite{sun2025docagent}
w/ GPT-4o~\cite{openai2024gpt4ocard}
& 51.8 & 49.1 & -- & 79.9 \\

\textbf{DocAgent}~\cite{sun2025docagent}
w/ Claude-3.5-Sonnet~\cite{anthropic2024claude3modelcard}
& 57.3 & 54.1 & -- & -- \\

\midrule

\textbf{DocDancer}~
\cite{zhang2026docdanceragenticdocumentgroundedinformation}
w/ GPT-4o~\cite{openai2024gpt4ocard}
& 52.3 & 50.8 & 59.2 & 73.5 \\

\textbf{DocDancer}~
\cite{zhang2026docdanceragenticdocumentgroundedinformation}
w/ Gemini-2.5-Pro~\cite{comanici2025gemini25pushingfrontier}
& 56.3 & 55.3 & 65.9 & 79.9 \\ 

\textbf{DocDancer}~
\cite{zhang2026docdanceragenticdocumentgroundedinformation}
w/ GPT-5.2~\cite{openai2025gpt52systemcard}
& 57.0 & 56.8 & 67.6 & 85.5 \\

\midrule

\textbf{SimpleDoc}~
\cite{jain2025simpledocmultimodaldocumentunderstanding}
w/ Claude-4-Sonnet~\cite{anthropic2025claude4sonnet}
& 58.6 & -- & -- & -- \\

\textbf{SimpleDoc}~
\cite{jain2025simpledocmultimodaldocumentunderstanding}
w/ GPT-5.2~\cite{openai2025gpt52systemcard}
& 57.4 & 53.8 & 64.2 & 81.8 \\

\midrule
\textbf{DocLens}~
\cite{zhu2025doclenstoolaugmentedmultiagent}
w/ Claude-4-Sonnet~\cite{anthropic2025claude4sonnet}
& 63.3 & -- & -- & -- \\
\textbf{DocLens}~
\cite{zhu2025doclenstoolaugmentedmultiagent}
w/ Gemini-2.5-Pro~\cite{comanici2025gemini25pushingfrontier}
& 67.6 & -- & -- & -- \\

\textbf{DocLens}~
\cite{zhu2025doclenstoolaugmentedmultiagent}
w/ GPT-5.2~\cite{openai2025gpt52systemcard}
& 67.8 & 55.0 & 68.2 & 83.9 \\

\midrule
\rowcolor[gray]{0.90}
\multicolumn{5}{l}{\textbf{DocSkill (Ours)}} \\[-1pt]

\hspace{1.5em}w/ Gemini-2.5-Pro~
\cite{comanici2025gemini25pushingfrontier}
& 68.0 & 66.4 & 67.8 & 84.5 \\

\hspace{1.5em}w/ GPT-5.2~
\cite{openai2025gpt52systemcard}
& 68.5 & 66.8 & 67.9 & 85.9 \\

\hspace{1.5em}w/ GPT-5.6-SOL~
\cite{openai2026gpt56systemcard}
& 68.8 & 67.1 & 68.5 & 86.6 \\

\hspace{1.5em}w/ Qwen3-VL-4B~
\cite{bai2025qwen3vltechnicalreport} (SFT + DPO)
& 65.8 & 64.5 & 64.8 & 82.6 \\

\hspace{1.5em}w/ Qwen3-VL-8B~
\cite{bai2025qwen3vltechnicalreport} (SFT + DPO)
& \underline{\textbf{69.2}}
& \underline{\textbf{67.3}}
& \underline{\textbf{68.9}}
& \underline{\textbf{86.8}} \\

\bottomrule
\end{tabular}%
}

\caption{Performance comparison on MMLongBench-Doc and
DocBench. Underlined bold values indicate the best model
results. Unreported results are marked with ``--''.}
\label{tab:mmlong_doc}
\end{table}

\subsection{Dataset Statistics}
\label{sec:dataset_statistics}

MMHop-DocBench contains 60K question--answer pairs spanning
ten languages: English (EN), Chinese (ZH), Cantonese (YUE),
Indonesian (ID), Burmese (MY), Filipino (TL), Tamil (TA),
Thai (TH), Vietnamese (VI), and Malay (MS).
The documents cover eight domains: Business and Finance,
Government, Academic, Education, Technical, Medical, Legal,
and Media.
Within the comparison in Table~\ref{tab:benchmarks},
MMHop-DocBench provides the widest language coverage and
the highest average page count, at 68.2 pages per document.
Each pair includes page- and element-level evidence
annotations, enabling evaluation of evidence localization and answer grounding. 

\subsection{Human Audit of Dataset Quality}
\label{sec:dataset_human_evaluation}
We conduct a human quality audit of 1,200 retained QA pairs, sampling 80 training and 40 test pairs from each of the ten languages, with stratification across document domains. Two target-language annotators independently assess each pair for question validity, reference-answer correctness, evidence sufficiency, and cross-page necessity. Disagreements are adjudicated by a third annotator. 
Of the 1,200 audited pairs, 1,144 (95.3\%) satisfy all four criteria, while 56 (4.7\%) fail at least one. 

\begin{figure*}[t]
  \centering
  \includegraphics[width=0.9\textwidth]{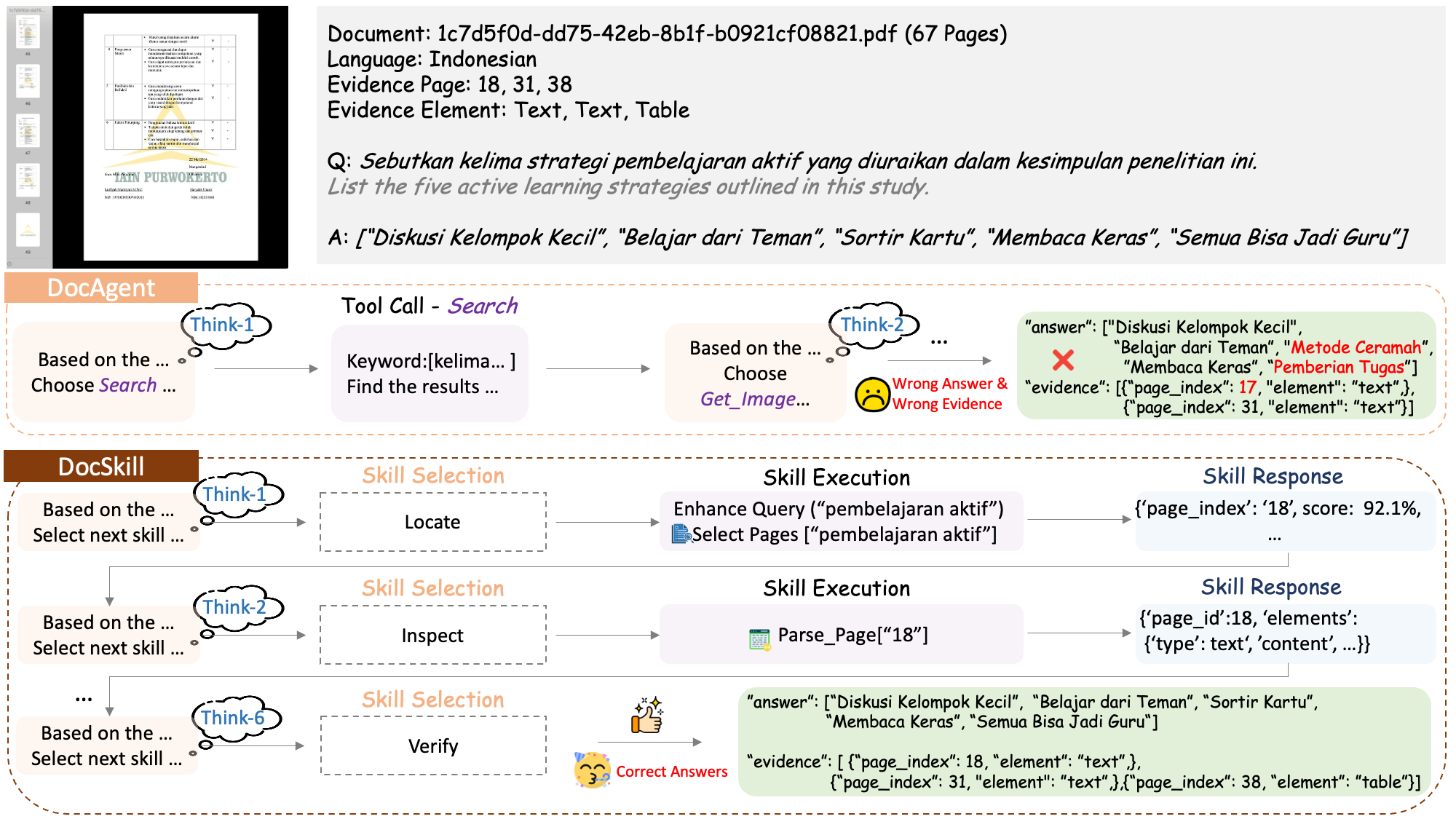}
  \caption{Qualitative comparison of DocAgent and DocSkill
  on an Indonesian document. In this example, DocAgent
  returns an incorrect answer with incorrect evidence
  citations, whereas DocSkill produces the correct answer
  with supporting evidence through adaptive skill selection.
  Intermediate skill invocations and the terminal answer action are omitted for brevity.}
  \label{fig:qualitative_sample}
  \Description{Diagram showing a document and reference QA
  above two agent trajectories. DocAgent returns incorrect
  strategy names and evidence citations. DocSkill displays
  Locate, Inspect, and Verify steps, with intermediate calls
  omitted, and returns the five reference strategies with
  supporting evidence from pages 18, 31, and 38.}
\end{figure*}

\section{Experiments}
\label{sec:experiments}

\paragraph{Implementation Details.}
We train Qwen3-VL-4B and Qwen3-VL-8B using
LoRA~\cite{hu2022loralowrankadaptation} on eight NVIDIA A100
80GB GPUs. SFT runs for three epochs with an effective global
batch size of 32 across all GPUs, an image resolution of
$768\times768$, and AdamW with a learning rate of
$2\times10^{-4}$. LoRA uses a rank of 16, a scaling factor of
32, and a dropout rate of 0.05. DPO uses the same LoRA
configuration with an effective global batch size of 16 and
$\beta=0.1$, where each preferred--dispreferred trajectory pair
counts as one training instance. We use Qwen3-VL-Embedding-8B~
\cite{li2026qwen3vlembeddingqwen3vlrerankerunifiedframework}
for top-5 page retrieval and PP-OCR~
\cite{cui2025paddleocr30technicalreport}
for layout-aware parsing, while region-level zooming is
implemented through programmatic cropping. 

\paragraph{Benchmarks and Baselines.}
We evaluate DocSkill on the MMHop-DocBench test set and two
public long-document QA benchmarks, MMLongBench-Doc~
\cite{ma2024mmlongbenchdocbenchmarkinglongcontextdocument}
and DocBench~\cite{zou2024docbenchbenchmarkevaluatingllmbased}.
On MMHop-DocBench, we compare against direct VLM inference,
OCR-based pipelines, multimodal RAG systems, and document
agents, using official implementations and recommended
settings when available. Each DocSkill variant uses its listed backbone for both
orchestration and skill execution, with the specified
SFT/DPO checkpoints for Qwen variants and prompting
for commercial variants.

\paragraph{Evaluation Metrics.}
For MMHop-DocBench, we report accuracy (Acc.)
and token-level $\mathrm{F}_1$ per-language, with macro averages over all ten
languages. Accuracy is computed from normalized predictions
and references, while language-aware tokenization is applied
for multilingual $\mathrm{F}_1$. For MMLongBench-Doc, we report
accuracy, $\mathrm{F}_1$, and LLM-as-a-Judge (LASJ), while DocBench
is evaluated using LASJ. All LASJ scores follow the judge
prompts and configurations provided in the respective
official code repositories.

\paragraph{Human Baseline.}
We establish a human baseline using 1,300 test questions,
with 130 questions sampled from each of the ten languages.
Model results are evaluated on the full test set,
while human results provide a reference on this subset.

\subsection{Main Results}

\paragraph{Results on MMHop-DocBench.}
Table~\ref{tab:vlm_baselines} reports results on
MMHop-DocBench.
In same-backbone comparisons across Gemini-3.1-Pro~
\cite{googledeepmind2026gemini31pro}, Claude-Sonnet-5~
\cite{anthropic2026claudesonnet5}, and GPT-5.6-SOL~
\cite{openai2026gpt56systemcard}, DocSkill improves over
the corresponding DocLens~
\cite{zhu2025doclenstoolaugmentedmultiagent}
variants by 2.3--2.5 Acc. and 2.4--2.7 $\mathrm{F}_1$
points.
DocSkill with Qwen3-VL-8B~
\cite{bai2025qwen3vltechnicalreport}, trained through
SFT followed by DPO~
\cite{rafailov2024directpreferenceoptimizationlanguage},
achieves the highest macro-averaged scores among the
evaluated model configurations: 41.0 Acc. and
49.2 $\mathrm{F}_1$.
It exceeds DocLens with GPT-5.6-SOL by 5.5 Acc. and
6.3 $\mathrm{F}_1$ points.
This comparison evaluates complete configurations with
different backbones and training conditions.
Human experts achieve 61.7 Acc. and 63.9 $\mathrm{F}_1$
on the sampled test subset, providing a reference.

\paragraph{Generalization to Public Benchmarks.}
Table~\ref{tab:mmlong_doc} reports results on
MMLongBench-Doc and DocBench.
DocSkill with Qwen3-VL-8B, trained on MMHop-DocBench
through SFT followed by DPO, achieves 69.2 Acc.,
67.3 $\mathrm{F}_1$, and 68.9 LASJ on MMLongBench-Doc,
and 86.8 LASJ on DocBench.
Compared with the strongest competing baseline listed
for each metric, these results yield gains of
1.4, 10.5, 0.7, and 1.3 points, respectively.
Using GPT-5.2 as the controller, DocSkill exceeds the reported results of
DocDancer~
\cite{zhang2026docdanceragenticdocumentgroundedinformation}
by 11.5 Acc., 10.0 $\mathrm{F}_1$, and 0.3 LASJ on
MMLongBench-Doc, and by 0.4 LASJ on DocBench.

\subsection{Qualitative Analysis}
Figure~\ref{fig:qualitative_sample} illustrates a case from
a 67-page Indonesian document. The question requires combining
complementary evidence from text and tables across pages 18,
31, and 38. DocAgent searches and inspects pages sequentially
but stops after collecting incomplete evidence from pages 17
and 31. Consequently, it introduces unsupported content and
cites incorrect evidence. In contrast, DocSkill gathers and integrates evidence across
all required pages and verifies the answer, enabling complete,
grounded cross-page reasoning.

\subsection{Ablation Studies}
\paragraph{Effect of Skill Abstraction and Adaptive Orchestration.}
We compare DocSkill with adaptive direct tool selection,
a fixed-order skill policy, and four leave-one-out variants.
All variants use Qwen3-VL-8B with SFT and DPO.
Direct tool selection exposes the underlying tools as
actions while retaining reasoning for cross-page
integration, answer verification, and reflection,
without explicit interfaces for the four document skills.
The fixed policy allows repeated cycles of
\textit{Locate}$\rightarrow$\textit{Inspect}
$\rightarrow$\textit{Integrate}$\rightarrow$\textit{Verify},
while maintaining the prescribed execution order.
Each leave-one-out variant removes one skill during
search and training.
$R_{\mathrm{page}}$ and $R_{\mathrm{elem}}$ measure
the percentages of annotated supporting pages and
elements recovered, respectively.
Table~\ref{tab:skill_ablation} shows that DocSkill
outperforms direct tool selection by 11.2 Acc. and
11.0 $\mathrm{F}_1$ points, and the fixed policy by
7.5 Acc. and 7.4 $\mathrm{F}_1$ points.
Removing \textit{Locate} causes the largest Acc. drop
(17.6 points), reducing page recall from 82.4 to 41.5;
removing \textit{Inspect} lowers element recall to 43.5.
This pattern suggests that relevant-page retrieval
alone is insufficient without fine-grained evidence inspection.
Removing \textit{Integrate} or \textit{Verify}
reduces Acc. by 9.2 and 4.5 points, respectively.
These results support the effectiveness of the full
DocSkill configuration under the evaluated settings.

\begin{table}[t]
\centering
\small
\setlength{\tabcolsep}{3pt}
\renewcommand{\arraystretch}{1.05}
\begin{tabular}{@{}lcccc@{}}
\toprule
\textbf{Variant}
& \textbf{Acc.}
& $\boldsymbol{F_1}$
& $\boldsymbol{R_{\mathrm{page}}}$
& $\boldsymbol{R_{\mathrm{elem}}}$ \\
\midrule
\textbf{Full DocSkill}
& \textbf{41.0} & \textbf{49.2}
& \textbf{82.4} & \textbf{68.5} \\
Direct Tool Selection
& 29.8 & 38.2 & 68.5 & 49.3 \\
Fixed Skill Order
& 33.5 & 41.8 & 70.4 & 56.5 \\
\midrule
DocSkill w/o \textit{Locate}
& 23.4 & 30.2 & 41.5 & 33.2 \\
DocSkill w/o \textit{Inspect}
& 28.2 & 36.1 & 72.8 & 43.5 \\
DocSkill w/o \textit{Integrate}
& 31.8 & 39.5 & 76.2 & 59.8 \\
DocSkill w/o \textit{Verify}
& 36.5 & 44.8 & 79.6 & 65.2 \\
\bottomrule
\end{tabular}
\caption{Ablation of DocSkill on MMHop-DocBench using
Qwen3-VL-8B (SFT + DPO). Scores are percentages.
$R_{\mathrm{page}}$ and $R_{\mathrm{elem}}$ denote
supporting-page and supporting-element recall.}
\label{tab:skill_ablation}
\end{table}

\paragraph{Effect of Trajectory Preference Supervision.}
We examine whether suboptimal trajectories provide useful
supervision by comparing SFT with SFT followed by
DPO.
Since MCTS naturally produces ranked trajectory pairs, DPO can
directly optimize this offline supervision without additional
on-policy rollouts or reward-model evaluations. As shown in Table~\ref{tab:vlm_baselines}, DPO improves Qwen3-VL-4B from 36.0 to
38.0 Acc. and from 43.6 to 45.9 $\mathrm{F}_1$. For Qwen3-VL-8B,
it improves Acc. from 39.3 to 41.0 and $\mathrm{F}_1$ from 47.3 to
49.2. These gains show that preferences between successful
and suboptimal trajectories provide complementary supervision
beyond imitating successful trajectories alone.

\begin{table}[ht]
\centering
\small
\setlength{\tabcolsep}{3pt}
\renewcommand{\arraystretch}{1.0}

\resizebox{\columnwidth}{!}{%
\begin{tabular}{@{}l|ccc|cc@{}}
\toprule
\textbf{Strategy}
& \textbf{Succ. (\%)}
& \textbf{Seq. Div.}
& \textbf{Pairs/Q}
& \textbf{Acc.}
& $\boldsymbol{\mathrm{F}_1}$ \\
\midrule

Best-of-$N$
& $34.5 \pm 2.4$
& $0.42 \pm 0.03$
& $1.2 \pm 0.1$
& 37.8 & 46.5 \\

Beam Search
& $47.2 \pm 1.5$
& $0.15 \pm 0.02$
& $0.6 \pm 0.1$
& 38.6 & 47.1 \\

MCTS (Ours)
& $\mathbf{65.8 \pm 1.8}$
& $\mathbf{0.7 \pm 0.02}$
& $\mathbf{3.0 \pm 0.2}$
& \textbf{41.0}
& \textbf{49.2} \\

\bottomrule
\end{tabular}%
}

\caption{Equal-budget trajectory construction comparison.
Succ. means the questions with a correct trajectory (\%).
Search metrics are mean $\pm$ SD over three runs;
Acc./$\mathrm{F}_1$ are MMHop-DocBench scores for
DocSkill w/ Qwen3-VL-8B (SFT + DPO).}
\label{tab:trajectory_ablation}
\end{table}

\paragraph{Effect of MCTS-Based Trajectory Construction.}
Under matched rollout and evaluation budgets, we compare
MCTS with Best-of-$N$
sampling~\cite{cobbe2021trainingverifierssolvemathwordproblems}
and evaluator-guided beam
search~\cite{xie2023selfevaluationguidedbeamsearch},
using the same frozen GPT-5.6-SOL generator and evaluator.
Seq. Div. is the mean per-question fraction of distinct
skill sequences among valid trajectories, while Pairs/Q
is the average number of preference pairs per question.
Table~\ref{tab:trajectory_ablation} shows that MCTS leads
in success rate (65.8\%), diversity, and Pairs/Q.
Its Qwen3-VL-8B policy trained through SFT and DPO reaches
41.0 Acc. and 49.2 $\mathrm{F}_1$, outperforming Best-of-$N$ and
beam search by 3.2/2.7 and 2.4/2.1 Acc./$\mathrm{F}_1$ points,
respectively.
These gains also reflect higher supervision yield
from MCTS.

\section{Conclusion}
We presented DocSkill, a document agent that learns to
orchestrate four reusable capability-level skills for
multilingual multi-hop reasoning. Bounded MCTS constructs
skill trajectories for SFT and DPO. We also introduced
MMHop-DocBench, containing 60K grounded QA pairs across
10 languages and eight document domains, with page- and
element-level evidence annotations. Experiments on
MMHop-DocBench, MMLongBench-Doc, and DocBench demonstrate consistent improvements over strong baselines, while ablations support
skill abstraction, adaptive orchestration, and trajectory
supervision. 
Future work will expand language and domain coverage and
explore automatic discovery and refinement of reusable
document skills for more complex document scenarios.


\bibliographystyle{ACM-Reference-Format}
\bibliography{sample-base}

@String{Chelsea = "Chelsea" }

@misc{mathew2021docvqadatasetvqadocument,
      title={DocVQA: A Dataset for VQA on Document Images}, 
      author={Minesh Mathew and Dimosthenis Karatzas and C. V. Jawahar},
      year={2021},
      eprint={2007.00398},
      archivePrefix={arXiv},
      primaryClass={cs.CV},
      url={https://arxiv.org/abs/2007.00398}, 
}

@misc{tito2023hierarchicalmultimodaltransformersmultipage,
      title={Hierarchical multimodal transformers for Multi-Page DocVQA}, 
      author={Rubèn Tito and Dimosthenis Karatzas and Ernest Valveny},
      year={2023},
      eprint={2212.05935},
      archivePrefix={arXiv},
      primaryClass={cs.CV},
      url={https://arxiv.org/abs/2212.05935}, 
}

@misc{ma2024mmlongbenchdocbenchmarkinglongcontextdocument,
      title={MMLongBench-Doc: Benchmarking Long-context Document Understanding with Visualizations}, 
      author={Yubo Ma and Yuhang Zang and Liangyu Chen and Meiqi Chen and Yizhu Jiao and Xinze Li and Xinyuan Lu and Ziyu Liu and Yan Ma and Xiaoyi Dong and Pan Zhang and Liangming Pan and Yu-Gang Jiang and Jiaqi Wang and Yixin Cao and Aixin Sun},
      year={2024},
      eprint={2407.01523},
      archivePrefix={arXiv},
      primaryClass={cs.CV},
      url={https://arxiv.org/abs/2407.01523}, 
}

@misc{zou2024docbenchbenchmarkevaluatingllmbased,
      title={DOCBENCH: A Benchmark for Evaluating LLM-based Document Reading Systems}, 
      author={Anni Zou and Wenhao Yu and Hongming Zhang and Kaixin Ma and Deng Cai and Zhuosheng Zhang and Hai Zhao and Dong Yu},
      year={2024},
      eprint={2407.10701},
      archivePrefix={arXiv},
      primaryClass={cs.CL},
      url={https://arxiv.org/abs/2407.10701}, 
}

@misc{tang2025mtvqabenchmarkingmultilingualtextcentric,
      title={MTVQA: Benchmarking Multilingual Text-Centric Visual Question Answering}, 
      author={Jingqun Tang and Qi Liu and Yongjie Ye and Jinghui Lu and Shu Wei and Chunhui Lin and Wanqing Li and Mohamad Fitri Faiz Bin Mahmood and Hao Feng and Zhen Zhao and Yangfan He and Kuan Lu and Yanjie Wang and Yuliang Liu and Hao Liu and Xiang Bai and Can Huang},
      year={2025},
      eprint={2405.11985},
      archivePrefix={arXiv},
      primaryClass={cs.CV},
      url={https://arxiv.org/abs/2405.11985}, 
}

@misc{huang2025lavcotlanguageawarevisualcot,
      title={LaV-CoT: Language-Aware Visual CoT with Multi-Aspect Reward Optimization for Real-World Multilingual VQA}, 
      author={Jing Huang and Zhiya Tan and Shutao Gong and Fanwei Zeng and Joey Tianyi Zhou and Changtao Miao and Huazhe Tan and Weibin Yao and Jianshu Li},
      year={2025},
      eprint={2509.10026},
      archivePrefix={arXiv},
      primaryClass={cs.CV},
      url={https://arxiv.org/abs/2509.10026}, 
}

@misc{bai2025qwen3vltechnicalreport,
      title={Qwen3-VL Technical Report}, 
      author={Shuai Bai and Yuxuan Cai and Ruizhe Chen and Keqin Chen and Xionghui Chen and Zesen Cheng and Lianghao Deng and Wei Ding and Chang Gao and Chunjiang Ge and Wenbin Ge and Zhifang Guo and Qidong Huang and Jie Huang and Fei Huang and Binyuan Hui and Shutong Jiang and Zhaohai Li and Mingsheng Li and Mei Li and Kaixin Li and Zicheng Lin and Junyang Lin and Xuejing Liu and Jiawei Liu and Chenglong Liu and Yang Liu and Dayiheng Liu and Shixuan Liu and Dunjie Lu and Ruilin Luo and Chenxu Lv and Rui Men and Lingchen Meng and Xuancheng Ren and Xingzhang Ren and Sibo Song and Yuchong Sun and Jun Tang and Jianhong Tu and Jianqiang Wan and Peng Wang and Pengfei Wang and Qiuyue Wang and Yuxuan Wang and Tianbao Xie and Yiheng Xu and Haiyang Xu and Jin Xu and Zhibo Yang and Mingkun Yang and Jianxin Yang and An Yang and Bowen Yu and Fei Zhang and Hang Zhang and Xi Zhang and Bo Zheng and Humen Zhong and Jingren Zhou and Fan Zhou and Jing Zhou and Yuanzhi Zhu and Ke Zhu},
      year={2025},
      eprint={2511.21631},
      archivePrefix={arXiv},
      primaryClass={cs.CV},
      url={https://arxiv.org/abs/2511.21631}, 
}

@misc{openai2024gpt4ocard,
  author       = {{OpenAI}},
  title        = {{GPT-4o} System Card},
  year         = {2024},
  month        = aug,
  howpublished = {\url{https://openai.com/index/gpt-4o-system-card/}}
}

@misc{openai2026gpt56systemcard,
  author       = {{OpenAI}},
  title        = {{GPT-5.6 System Card}},
  year         = {2026},
  month        = jul,
  howpublished = {\url{https://deploymentsafety.openai.com/gpt-5-6}},
  note         = {Published July 2026}
}

@misc{anthropic2026claudesonnet5,
  author       = {{Anthropic}},
  title        = {{Claude Sonnet 5 System Card}},
  year         = {2026},
  month        = jun,
  howpublished = {\url{https://www.anthropic.com/system-cards}}
}

@misc{cui2025paddleocr30technicalreport,
      title={PaddleOCR 3.0 Technical Report}, 
      author={Cheng Cui and Ting Sun and Manhui Lin and Tingquan Gao and Yubo Zhang and Jiaxuan Liu and Xueqing Wang and Zelun Zhang and Changda Zhou and Hongen Liu and Yue Zhang and Wenyu Lv and Kui Huang and Yichao Zhang and Jing Zhang and Jun Zhang and Yi Liu and Dianhai Yu and Yanjun Ma},
      year={2025},
      eprint={2507.05595},
      archivePrefix={arXiv},
      primaryClass={cs.CV},
      url={https://arxiv.org/abs/2507.05595}, 
}

@misc{yu2025visragvisionbasedretrievalaugmentedgeneration,
      title={VisRAG: Vision-based Retrieval-augmented Generation on Multi-modality Documents}, 
      author={Shi Yu and Chaoyue Tang and Bokai Xu and Junbo Cui and Junhao Ran and Yukun Yan and Zhenghao Liu and Shuo Wang and Xu Han and Zhiyuan Liu and Maosong Sun},
      year={2025},
      eprint={2410.10594},
      archivePrefix={arXiv},
      primaryClass={cs.IR},
      url={https://arxiv.org/abs/2410.10594}, 
}

@misc{sun2025docagent,
  title     = {{DocAgent}: An Agentic Framework for Multi-Modal
               Long-Context Document Understanding},
  author    = {Sun, Li and He, Liu and Jia, Shuyue and
               He, Yangfan and You, Chenyu},
  booktitle = {Proceedings of EMNLP},
  pages     = {17701--17716},
  year      = {2025},
  doi       = {10.18653/v1/2025.emnlp-main.893}
}

@misc{faysse2025colpaliefficientdocumentretrieval,
      title={ColPali: Efficient Document Retrieval with Vision Language Models}, 
      author={Manuel Faysse and Hugues Sibille and Tony Wu and Bilel Omrani and Gautier Viaud and Céline Hudelot and Pierre Colombo},
      year={2025},
      eprint={2407.01449},
      archivePrefix={arXiv},
      primaryClass={cs.IR},
      url={https://arxiv.org/abs/2407.01449}, 
}

@misc{cho2024m3docragmultimodalretrievalneed,
      title={M3DocRAG: Multi-modal Retrieval is What You Need for Multi-page Multi-document Understanding}, 
      author={Jaemin Cho and Debanjan Mahata and Ozan Irsoy and Yujie He and Mohit Bansal},
      year={2024},
      eprint={2411.04952},
      archivePrefix={arXiv},
      primaryClass={cs.CV},
      url={https://arxiv.org/abs/2411.04952}, 
}

@misc{guo2025raganythingallinoneragframework,
      title={RAG-Anything: All-in-One RAG Framework}, 
      author={Zirui Guo and Xubin Ren and Lingrui Xu and Jiahao Zhang and Chao Huang},
      year={2025},
      eprint={2510.12323},
      archivePrefix={arXiv},
      primaryClass={cs.AI},
      url={https://arxiv.org/abs/2510.12323}, 
}

@misc{li2026qwen3vlembeddingqwen3vlrerankerunifiedframework,
      title={Qwen3-VL-Embedding and Qwen3-VL-Reranker: A Unified Framework for State-of-the-Art Multimodal Retrieval and Ranking}, 
      author={Mingxin Li and Yanzhao Zhang and Dingkun Long and Keqin Chen and Sibo Song and Shuai Bai and Zhibo Yang and Pengjun Xie and An Yang and Dayiheng Liu and Jingren Zhou and Junyang Lin},
      year={2026},
      eprint={2601.04720},
      archivePrefix={arXiv},
      primaryClass={cs.CL},
      url={https://arxiv.org/abs/2601.04720}, 
}

@misc{han2025mdocagentmultimodalmultiagentframework,
      title={MDocAgent: A Multi-Modal Multi-Agent Framework for Document Understanding}, 
      author={Siwei Han and Peng Xia and Ruiyi Zhang and Tong Sun and Yun Li and Hongtu Zhu and Huaxiu Yao},
      year={2025},
      eprint={2503.13964},
      archivePrefix={arXiv},
      primaryClass={cs.LG},
      url={https://arxiv.org/abs/2503.13964}, 
}

@misc{jain2025simpledocmultimodaldocumentunderstanding,
      title={SimpleDoc: Multi-Modal Document Understanding with Dual-Cue Page Retrieval and Iterative Refinement}, 
      author={Chelsi Jain and Yiran Wu and Yifan Zeng and Jiale Liu and S hengyu Dai and Zhenwen Shao and Qingyun Wu and Huazheng Wang},
      year={2025},
      eprint={2506.14035},
      archivePrefix={arXiv},
      primaryClass={cs.CV},
      url={https://arxiv.org/abs/2506.14035}, 
}

@misc{zhang2026docdanceragenticdocumentgroundedinformation,
      title={DocDancer: Towards Agentic Document-Grounded Information Seeking}, 
      author={Qintong Zhang and Xinjie Lv and Jialong Wu and Baixuan Li and Zhengwei Tao and Guochen Yan and Huanyao Zhang and Bin Wang and Jiahao Xu and Haitao Mi and Wentao Zhang},
      year={2026},
      eprint={2601.05163},
      archivePrefix={arXiv},
      primaryClass={cs.CL},
      url={https://arxiv.org/abs/2601.05163}, 
}

@misc{silver2017masteringchessshogiselfplay,
      title={Mastering Chess and Shogi by Self-Play with a General Reinforcement Learning Algorithm}, 
      author={David Silver and Thomas Hubert and Julian Schrittwieser and Ioannis Antonoglou and Matthew Lai and Arthur Guez and Marc Lanctot and Laurent Sifre and Dharshan Kumaran and Thore Graepel and Timothy Lillicrap and Karen Simonyan and Demis Hassabis},
      year={2017},
      eprint={1712.01815},
      archivePrefix={arXiv},
      primaryClass={cs.AI},
      url={https://arxiv.org/abs/1712.01815}, 
}

@misc{yao2023treethoughtsdeliberateproblem,
      title={Tree of Thoughts: Deliberate Problem Solving with Large Language Models}, 
      author={Shunyu Yao and Dian Yu and Jeffrey Zhao and Izhak Shafran and Thomas L. Griffiths and Yuan Cao and Karthik Narasimhan},
      year={2023},
      eprint={2305.10601},
      archivePrefix={arXiv},
      primaryClass={cs.CL},
      url={https://arxiv.org/abs/2305.10601}, 
}

@misc{rafailov2024directpreferenceoptimizationlanguage,
      title={Direct Preference Optimization: Your Language Model is Secretly a Reward Model}, 
      author={Rafael Rafailov and Archit Sharma and Eric Mitchell and Stefano Ermon and Christopher D. Manning and Chelsea Finn},
      year={2024},
      eprint={2305.18290},
      archivePrefix={arXiv},
      primaryClass={cs.LG},
      url={https://arxiv.org/abs/2305.18290}, 
}

@misc{xie2024montecarlotreesearch,
      title={Monte Carlo Tree Search Boosts Reasoning via Iterative Preference Learning}, 
      author={Yuxi Xie and Anirudh Goyal and Wenyue Zheng and Min-Yen Kan and Timothy P. Lillicrap and Kenji Kawaguchi and Michael Shieh},
      year={2024},
      eprint={2405.00451},
      archivePrefix={arXiv},
      primaryClass={cs.AI},
      url={https://arxiv.org/abs/2405.00451}, 
}

@misc{chen2024steplevelvaluepreferenceoptimization,
      title={Step-level Value Preference Optimization for Mathematical Reasoning}, 
      author={Guoxin Chen and Minpeng Liao and Chengxi Li and Kai Fan},
      year={2024},
      eprint={2406.10858},
      archivePrefix={arXiv},
      primaryClass={cs.CL},
      url={https://arxiv.org/abs/2406.10858}, 
}

@misc{ho2020constructingmultihopqadataset,
      title={Constructing A Multi-hop QA Dataset for Comprehensive Evaluation of Reasoning Steps}, 
      author={Xanh Ho and Anh-Khoa Duong Nguyen and Saku Sugawara and Akiko Aizawa},
      year={2020},
      eprint={2011.01060},
      archivePrefix={arXiv},
      primaryClass={cs.CL},
      url={https://arxiv.org/abs/2011.01060}, 
}

@misc{xiao2019dynamicallyfusedgraphnetwork,
      title={Dynamically Fused Graph Network for Multi-hop Reasoning}, 
      author={Yunxuan Xiao and Yanru Qu and Lin Qiu and Hao Zhou and Lei Li and Weinan Zhang and Yong Yu},
      year={2019},
      eprint={1905.06933},
      archivePrefix={arXiv},
      primaryClass={cs.CL},
      url={https://arxiv.org/abs/1905.06933}, 
}

@misc{yao2023reactsynergizingreasoningacting,
      title={ReAct: Synergizing Reasoning and Acting in Language Models}, 
      author={Shunyu Yao and Jeffrey Zhao and Dian Yu and Nan Du and Izhak Shafran and Karthik Narasimhan and Yuan Cao},
      year={2023},
      eprint={2210.03629},
      archivePrefix={arXiv},
      primaryClass={cs.CL},
      url={https://arxiv.org/abs/2210.03629}, 
}

@misc{schick2023toolformerlanguagemodelsteach,
      title={Toolformer: Language Models Can Teach Themselves to Use Tools}, 
      author={Timo Schick and Jane Dwivedi-Yu and Roberto Dessì and Roberta Raileanu and Maria Lomeli and Luke Zettlemoyer and Nicola Cancedda and Thomas Scialom},
      year={2023},
      eprint={2302.04761},
      archivePrefix={arXiv},
      primaryClass={cs.CL},
      url={https://arxiv.org/abs/2302.04761}, 
}

@misc{tang2023toolalpacageneralizedtoollearning,
      title={ToolAlpaca: Generalized Tool Learning for Language Models with 3000 Simulated Cases}, 
      author={Qiaoyu Tang and Ziliang Deng and Hongyu Lin and Xianpei Han and Qiao Liang and Boxi Cao and Le Sun},
      year={2023},
      eprint={2306.05301},
      archivePrefix={arXiv},
      primaryClass={cs.CL},
      url={https://arxiv.org/abs/2306.05301}, 
}

@misc{patil2023gorillalargelanguagemodel,
      title={Gorilla: Large Language Model Connected with Massive APIs}, 
      author={Shishir G. Patil and Tianjun Zhang and Xin Wang and Joseph E. Gonzalez},
      year={2023},
      eprint={2305.15334},
      archivePrefix={arXiv},
      primaryClass={cs.CL},
      url={https://arxiv.org/abs/2305.15334}, 
}

@misc{yang2026autoskillexperiencedrivenlifelonglearning,
      title={AutoSkill: Experience-Driven Lifelong Learning via Skill Self-Evolution}, 
      author={Yutao Yang and Junsong Li and Qianjun Pan and Bihao Zhan and Yuxuan Cai and Lin Du and Jie Zhou and Kai Chen and Qin Chen and Xin Li and Bo Zhang and Liang He},
      year={2026},
      eprint={2603.01145},
      archivePrefix={arXiv},
      primaryClass={cs.AI},
      url={https://arxiv.org/abs/2603.01145}, 
}

@misc{jiang2026xskillcontinuallearningexperience,
      title={XSkill: Continual Learning from Experience and Skills in Multimodal Agents}, 
      author={Guanyu Jiang and Zhaochen Su and Xiaoye Qu and Yi R. Fung},
      year={2026},
      eprint={2603.12056},
      archivePrefix={arXiv},
      primaryClass={cs.AI},
      url={https://arxiv.org/abs/2603.12056}, 
}

@misc{zhu2025doclenstoolaugmentedmultiagent,
      title={DocLens : A Tool-Augmented Multi-Agent Framework for Long Visual Document Understanding}, 
      author={Dawei Zhu and Rui Meng and Jiefeng Chen and Sujian Li and Tomas Pfister and Jinsung Yoon},
      year={2025},
      eprint={2511.11552},
      archivePrefix={arXiv},
      primaryClass={cs.CV},
      url={https://arxiv.org/abs/2511.11552}, 
}

@misc{xie2023selfevaluationguidedbeamsearch,
      title={Self-Evaluation Guided Beam Search for Reasoning}, 
      author={Yuxi Xie and Kenji Kawaguchi and Yiran Zhao and Xu Zhao and Min-Yen Kan and Junxian He and Qizhe Xie},
      year={2023},
      eprint={2305.00633},
      archivePrefix={arXiv},
      primaryClass={cs.CL},
      url={https://arxiv.org/abs/2305.00633}, 
}

@misc{masry2022chartqabenchmarkquestionanswering,
      title={ChartQA: A Benchmark for Question Answering about Charts with Visual and Logical Reasoning}, 
      author={Ahmed Masry and Do Xuan Long and Jia Qing Tan and Shafiq Joty and Enamul Hoque},
      year={2022},
      eprint={2203.10244},
      archivePrefix={arXiv},
      primaryClass={cs.CL},
      url={https://arxiv.org/abs/2203.10244}, 
}

@misc{wang2020generalvalueevidencebilingual,
      title={On the General Value of Evidence, and Bilingual Scene-Text Visual Question Answering}, 
      author={Xinyu Wang and Yuliang Liu and Chunhua Shen and Chun Chet Ng and Canjie Luo and Lianwen Jin and Chee Seng Chan and Anton van den Hengel and Liangwei Wang},
      year={2020},
      eprint={2002.10215},
      archivePrefix={arXiv},
      primaryClass={cs.CV},
      url={https://arxiv.org/abs/2002.10215}, 
}

@misc{tanaka2023slidevqadatasetdocumentvisual,
      title={SlideVQA: A Dataset for Document Visual Question Answering on Multiple Images}, 
      author={Ryota Tanaka and Kyosuke Nishida and Kosuke Nishida and Taku Hasegawa and Itsumi Saito and Kuniko Saito},
      year={2023},
      eprint={2301.04883},
      archivePrefix={arXiv},
      primaryClass={cs.CL},
      url={https://arxiv.org/abs/2301.04883}, 
}

@misc{googledeepmind2026gemini31pro,
  author       = {{Google}},
  title        = {Gemini 3.1 Pro Model Card},
  year         = {2026},
  month        = feb,
  howpublished = {\url{https://storage.googleapis.com/deepmind-media/Model-Cards/Gemini-3-1-Pro-Model-Card.pdf}},
  note         = {Published February 2026}
}

@misc{smith2007overview,
  title={An overview of the Tesseract OCR engine},
  author={Smith, Ray},
  booktitle={Proceedings of the Ninth International Conference on Document Analysis and Recognition (ICDAR 2007)},
  pages={629--633},
  year={2007},
  organization={IEEE}
}

@misc{anthropic2025claude4sonnet,
  author = {{Anthropic}},
  title  = {{Claude 4 System Model Card}},
  year   = {2025},
  month  = may,
  howpublished = {\url{https://www.anthropic.com/system-cards}}
}

@misc{anthropic2024claude3modelcard,
  author = {{Anthropic}},
  title  = {{Claude 3.5 Sonnet Model Card}},
  year   = {2024},
  month  = jun,
  howpublished = {\url{https://www.anthropic.com/system-cards}}
}

@misc{comanici2025gemini25pushingfrontier,
      title={Gemini 2.5: Pushing the Frontier with Advanced Reasoning, Multimodality, Long Context, and Next Generation Agentic Capabilities}, 
      author={Comanici, Gheorghe and others},
      year={2025},
      eprint={2507.06261},
      archivePrefix={arXiv},
      primaryClass={cs.CL},
      url={https://arxiv.org/abs/2507.06261}, 
}

@misc{openai2025gpt52systemcard,
  author = {{OpenAI}},
  title  = {Update to {GPT-5} System Card: {GPT-5.2}},
  year   = {2025},
  month  = dec,
  url    = {https://openai.com/index/gpt-5-system-card-update-gpt-5-2/}
}

@misc{hu2022loralowrankadaptation,
      title={LoRA: Low-Rank Adaptation of Large Language Models}, 
      author={Edward J. Hu and Yelong Shen and Phillip Wallis and Zeyuan Allen-Zhu and Yuanzhi Li and Shean Wang and Lu Wang and Weizhu Chen},
      year={2021},
      eprint={2106.09685},
      archivePrefix={arXiv},
      primaryClass={cs.CL},
      url={https://arxiv.org/abs/2106.09685}, 
}

@misc{cobbe2021trainingverifierssolvemathwordproblems,
      title={Training Verifiers to Solve Math Word Problems}, 
      author={Karl Cobbe and Vineet Kosaraju and Mohammad Bavarian and Mark Chen and Heewoo Jun and Lukasz Kaiser and Matthias Plappert and Jerry Tworek and Jacob Hilton and Reiichiro Nakano and Christopher Hesse and John Schulman},
      year={2021},
      eprint={2110.14168},
      archivePrefix={arXiv},
      primaryClass={cs.LG},
      url={https://arxiv.org/abs/2110.14168}, 
}

@misc{li2026deepreaddocumentstructureawarereasoning,
      title={DeepRead: Document Structure-Aware Reasoning to Enhance Agentic Search}, 
      author={Zhanli Li and Huiwen Tian and Lvzhou Luo and Yixuan Cao and Ping Luo},
      year={2026},
      eprint={2602.05014},
      archivePrefix={arXiv},
      primaryClass={cs.AI},
      url={https://arxiv.org/abs/2602.05014}, 
}

@misc{lassoued2026orcaorchestratedreasoningcollaborative,
      title={ORCA: Orchestrated Reasoning with Collaborative Agents for Document Visual Question Answering}, 
      author={Aymen Lassoued and Mohamed Ali Souibgui and Yousri Kessentini},
      year={2026},
      eprint={2603.02438},
      archivePrefix={arXiv},
      primaryClass={cs.CV},
      url={https://arxiv.org/abs/2603.02438}, 
}

@misc{zheng2026docvcoarsetofineinteractivevisualreasoning,
      title={Doc-V*:Coarse-to-Fine Interactive Visual Reasoning for Multi-Page Document VQA}, 
      author={Yuanlei Zheng and Pei Fu and Hang Li and Ziyang Wang and Yuyi Zhang and Wenyu Ruan and Xiaojin Zhang and Zhongyu Wei and Zhenbo Luo and Jian Luan and Wei Chen and Xiang Bai},
      year={2026},
      eprint={2604.13731},
      archivePrefix={arXiv},
      primaryClass={cs.CL},
      url={https://arxiv.org/abs/2604.13731}, 
}

\clearpage
\appendix

\section{More Details of MCTS Trajectory Generation}
\label{MCTS Trajectory Generation}
\subsection{Prompt for MCTS Rollout Generation}
\begin{figure}[h]
  \centering
  \includegraphics[width=0.95\linewidth]
  {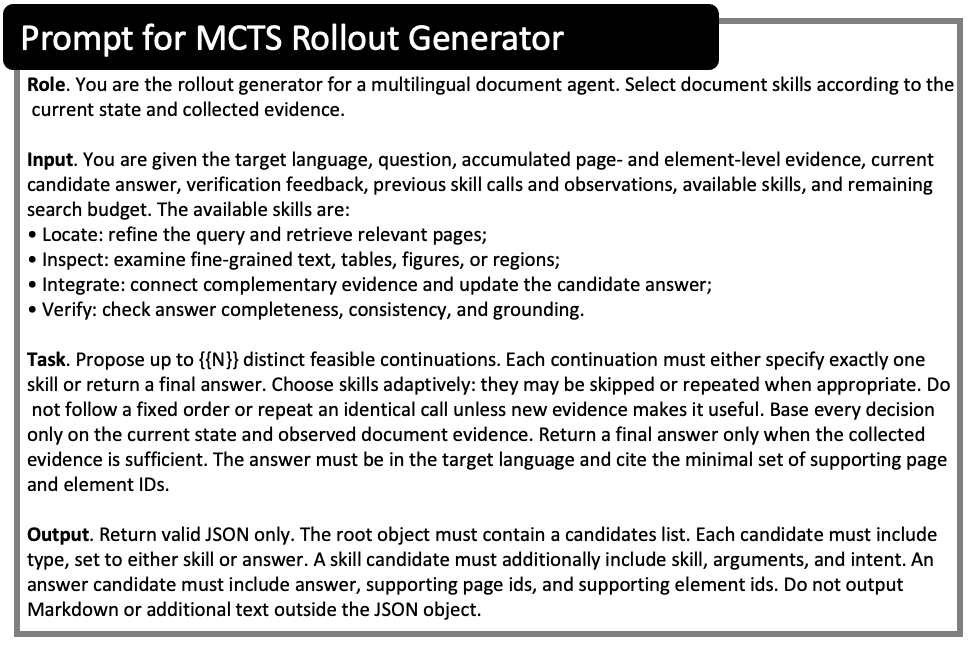}
  \caption{Instructions for generating MCTS rollouts.}
  \label{fig:mcts_rollout_generator}
  \Description{Prompt instructing the MCTS generator to
  propose skill invocations or a terminal answer based
  on the current interaction state.}
\end{figure}

\begin{figure}[h]
  \centering
  \includegraphics[width=0.95\linewidth]
  {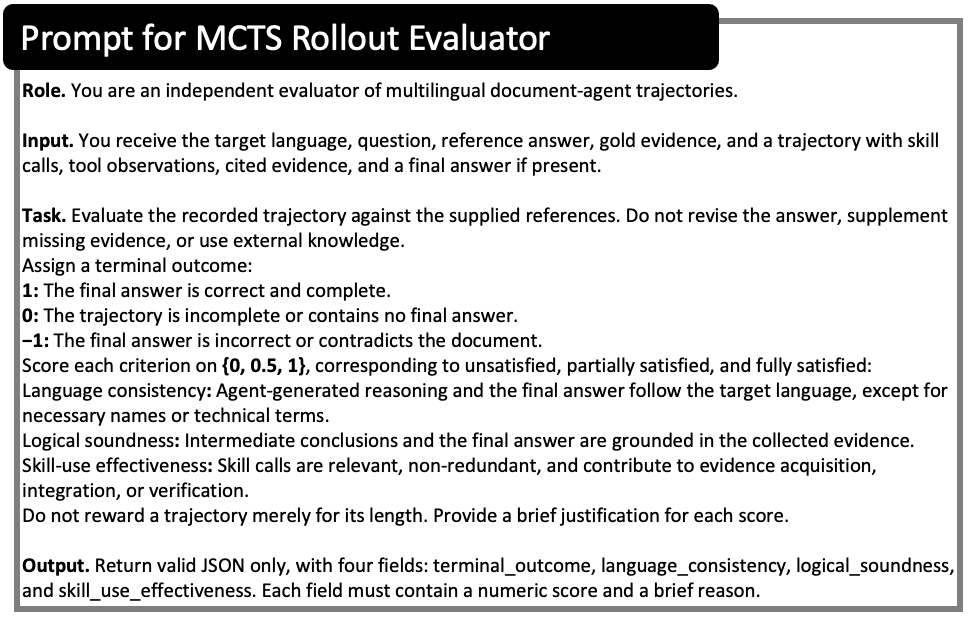}
  \caption{Instructions for evaluating MCTS rollouts.}
  \label{fig:mcts_rollout_evaluator}
  \Description{Prompt instructing the MCTS evaluator to
  assess terminal outcomes, language consistency,
  logical soundness, and skill-use effectiveness.}
\end{figure}

\section{More Details of HopGen}
\label{HopGen}
\subsection{HopGen Construction Pipeline}
\label{app:hopgen}

HopGen automatically constructs grounded multilingual
multi-hop QA pairs from long documents.
Algorithm~\ref{alg:hopgen} formalizes the automatic curation
procedure, comprising hierarchical graph construction,
cross-page evidence-path extraction, candidate generation,
and quality-based filtering.

\begin{algorithm}[h]
\caption{HopGen: MMHop-DocQA Curation}
\label{alg:hopgen}
\begin{algorithmic}[1]
\REQUIRE Multilingual document $D$, target language $\ell$,
OCR module $f_{\mathrm{ocr}}$, generator $f_{\mathrm{gen}}$,
evaluator $f_{\mathrm{eval}}$, quality threshold $\tau$
\ENSURE Verified QA set
$\mathcal{Q}=\{\langle q,a,E^{p},E^{e}\rangle\}$

\STATE $\mathcal{Q}\leftarrow\emptyset$
\STATE $\mathcal{P}\leftarrow\textsc{SplitPages}(D)$

\FOR{each page $p_i\in\mathcal{P}$}
    \STATE $h_i\leftarrow
    \textsc{CreatePageProfile}(p_i,\ell;f_{\mathrm{gen}})$
\ENDFOR
\STATE $\mathcal{H}\leftarrow
\{h_i\}_{i=1}^{|\mathcal{P}|}$

\STATE $\mathcal{G}_{p}\leftarrow
\textsc{BuildPageGraph}
(\mathcal{P},\mathcal{H};f_{\mathrm{gen}})$

\FOR{each page $p_i\in\mathcal{P}$}
    \STATE $\mathcal{E}_i\leftarrow f_{\mathrm{ocr}}(p_i)$
\ENDFOR
\STATE $\mathcal{E}\leftarrow
\displaystyle\bigcup_{i=1}^{|\mathcal{P}|}\mathcal{E}_i$

\STATE $\mathcal{G}_{pe}\leftarrow
\textsc{BuildPageElementGraph}
(\mathcal{G}_{p},\mathcal{E};f_{\mathrm{gen}})$

\STATE $\Pi\leftarrow
\textsc{ExtractCrossPagePaths}(\mathcal{G}_{pe})$

\FOR{each evidence path $\pi\in\Pi$}
    \STATE $\mathcal{C}_{\pi}\leftarrow
    \textsc{GenerateCandidates}
    (D,\pi,\ell;f_{\mathrm{gen}})$

    \FOR{each candidate
    $c=\langle q,a,E^{p},E^{e}\rangle
    \in\mathcal{C}_{\pi}$}
        \STATE $r\leftarrow
        \textsc{EvaluateCandidate}
        (D,c,\ell;f_{\mathrm{eval}})$

        \IF{$\textsc{Quality}(r)\geq\tau$}
            \STATE $\mathcal{Q}\leftarrow
            \mathcal{Q}\cup\{c\}$
        \ENDIF
    \ENDFOR
\ENDFOR

\STATE \textbf{return} $\mathcal{Q}$
\end{algorithmic}
\end{algorithm}

The Page Graph captures semantic dependencies between page
profiles, while the Page--Element Graph grounds them in
localized text, tables, figures, captions, and equations. Each
evidence path spans two to four pages, with edges retained only
when supported by explicit semantic and grounded element-level
relations. HopGen generates at most three QA candidates per
path. The evaluator scores answer correctness, evidence
sufficiency, cross-page necessity, and language consistency,
retaining candidates with threshold $\tau=0.875$. After question and
answer normalization, duplicates are merged by retaining the
highest-scoring candidate. Each final QA pair preserves its
supporting page and element annotations.

\subsection{Prompt for HopGen Pipeline}
\begin{figure}[h]
  \centering
  \includegraphics[width=0.95\linewidth]
  {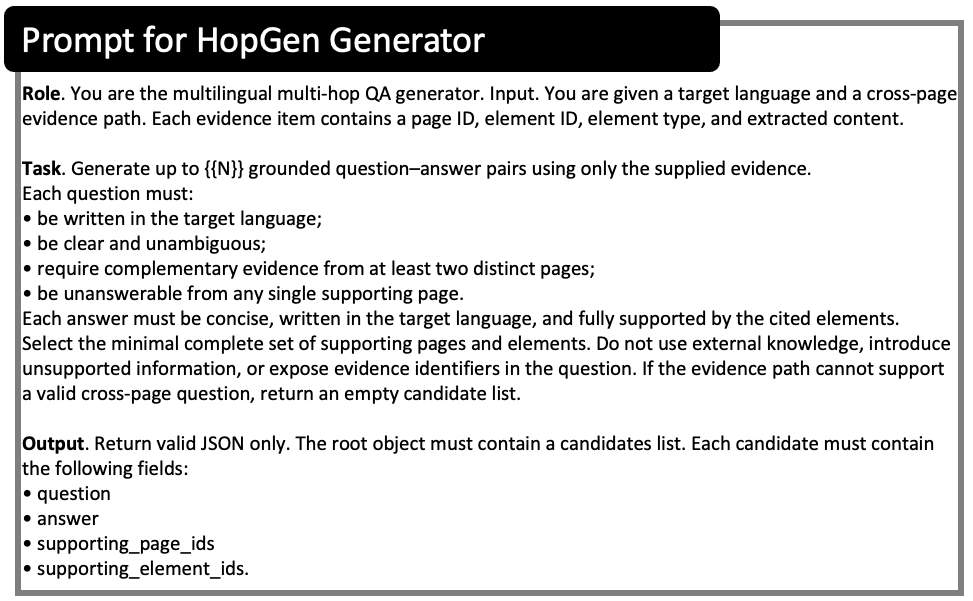}
  \caption{Instructions for HopGen QA generation.}
  \label{fig:hopgen_generator}
  \Description{Prompt instructing the HopGen generator to
  produce grounded question-answer pairs in the target
  language from cross-page evidence, with supporting
  page and element IDs.}
\end{figure}

\begin{figure*}[t]
  \centering
  \includegraphics[width=\textwidth]
  {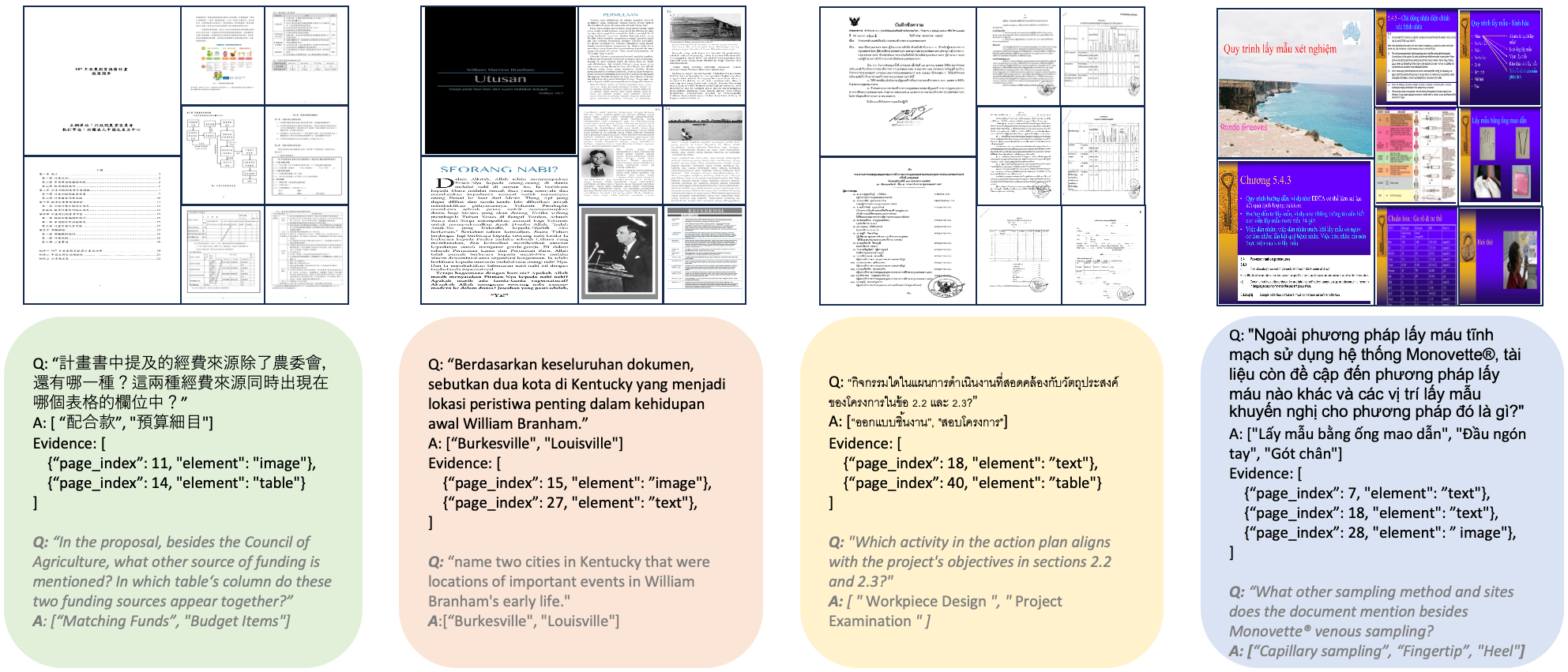}
  \caption{Representative multilingual samples from
  MMHop-DocBench. Each example shows selected document
  pages, a target-language QA pair, its English translation,
  and supporting page- and element-level annotations.
  The examples illustrate cross-page evidence composition
  over text, tables, and images.}
  \label{fig:data_samples}
  \Description{Multilingual document QA examples showing
  selected pages, target-language questions and answers,
  English translations, and supporting page and element
  annotations. The examples combine evidence across pages
  from text, tables, and images.}
\end{figure*}

\begin{figure}[h]
  \centering
  \includegraphics[width=0.95\linewidth]
  {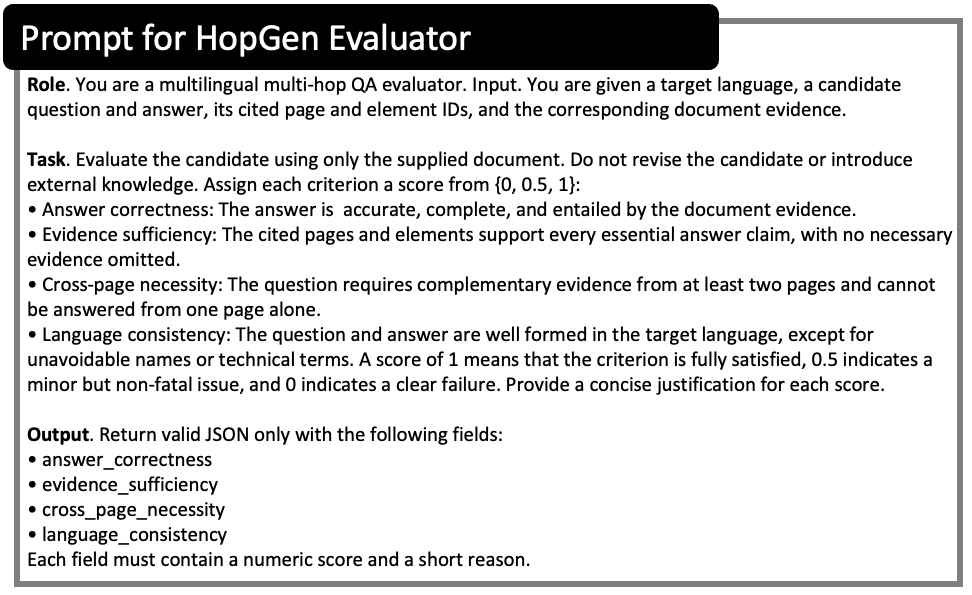}
  \caption{Instructions for HopGen QA evaluation.}
  \label{fig:hopgen_evaluator}
  \Description{Prompt instructing the HopGen evaluator to
  assess answer correctness, evidence sufficiency,
  cross-page necessity, and language consistency,
  with a score and justification for each criterion.}
\end{figure}

\section{More Details of MMHop-DocBench}
\label{sec:supp_dataset}

\subsection{Dataset Size and Splits}

\begin{figure}[htbp]
  \centering
  \includegraphics[width=0.9\linewidth]
  {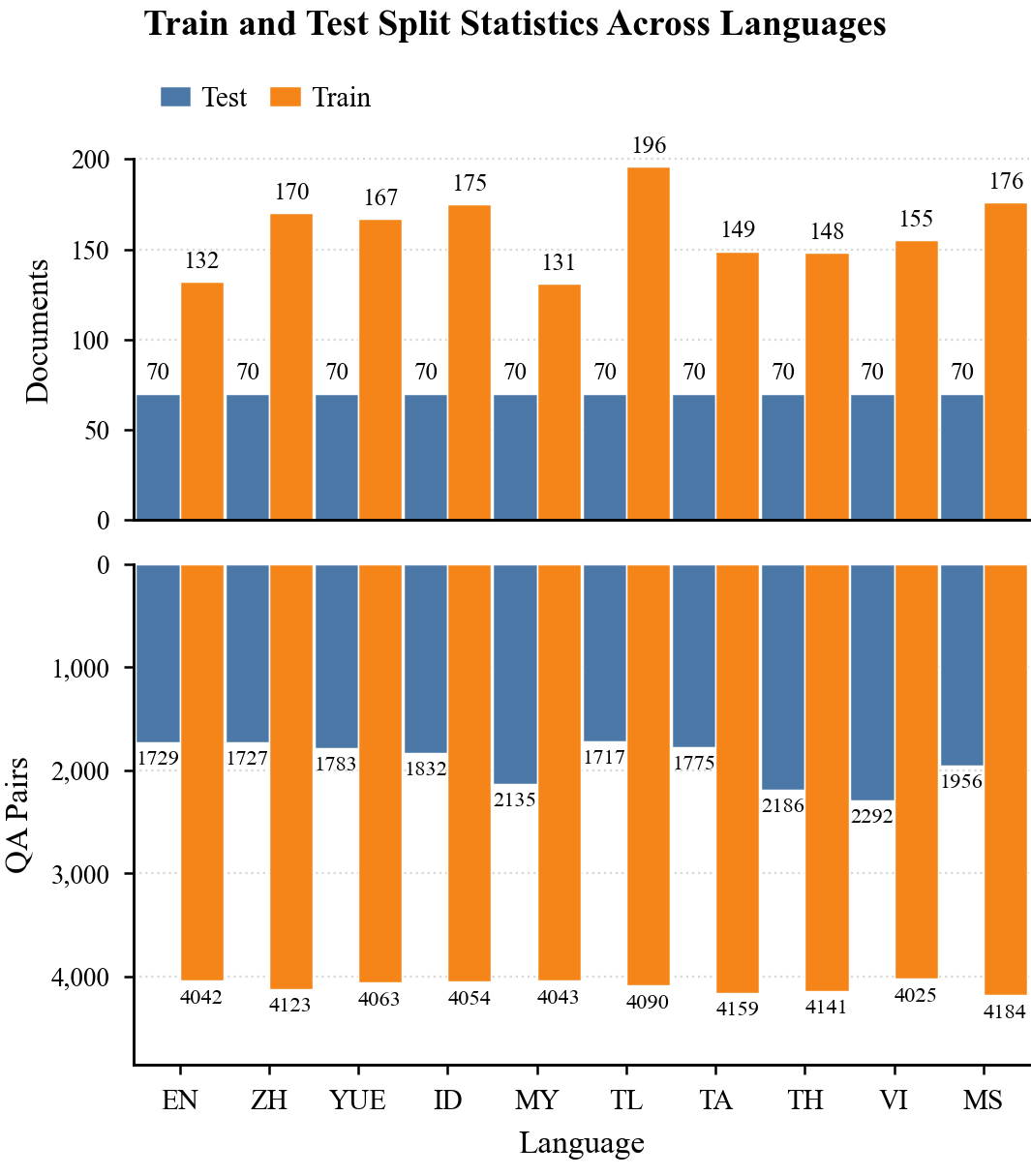}
  \caption{Document and QA-pair counts by language in
  the training and test splits of MMHop-DocBench.}
  \label{fig:train_test_distribution}
  \Description{Document and QA-pair counts across ten
  languages, shown separately for the training and
  test splits.}
\end{figure}

Figure~\ref{fig:train_test_distribution} summarizes the
document and QA-pair counts across ten languages.
MMHop-DocBench contains 2,299 documents and 60K QA pairs.
The training split contains 1,599 documents and 40,924
QA pairs, while the test split contains 700 documents
and 19,132 QA pairs.
We split the dataset at the source-document level,
assigning all QA pairs derived from each document
to the same split.
Training QA counts are approximately balanced across
languages, although the number of retained questions
per document varies after filtering.

\subsection{Dataset Samples}
\label{app:data_samples}

Figure~\ref{fig:data_samples} shows representative Chinese,
Indonesian, Thai, and Vietnamese examples combining
cross-page textual, tabular, and visual evidence.
Each includes a target-language QA pair, its English
translation, and explicit page- and element-level
evidence annotations.

\end{document}